\documentclass[a4paper,fleqn]{cas-sc}
\usepackage[authoryear]{natbib}
\usepackage{adjustbox}
\usepackage{multirow}
\usepackage{longtable}
\usepackage{geometry} % Adjust page margins if needed
\usepackage{tabularx}
\usepackage{booktabs}
\usepackage{xcolor}
\usepackage{lscape}
\usepackage{hyperref}
\usepackage{tabularx}
\usepackage{array}
\usepackage{amsmath}
\usepackage{amssymb}
\usepackage{cleveref}
\usepackage{adjustbox}
\usepackage{multirow}
\usepackage{longtable}
\usepackage{booktabs}
\usepackage{subcaption}
\usepackage{float}
\usepackage{algorithm}
\usepackage{algpseudocode}
\usepackage{graphicx}
\usepackage{epsfig}

\def\tsc#1{\csdef{#1}{\textsc{\lowercase{#1}}\xspace}}
\tsc{WGM}
\tsc{QE}
\tsc{EP}
\tsc{PMS}
\tsc{BEC}
\tsc{DE}
\begin{document}
\let\WriteBookmarks\relax
\def\floatpagepagefraction{1}
\def\textpagefraction{.001}
% \begin{frontmatter}
\shorttitle{ Fine-Tuned Offline RL Augmented Prescriptive Process Monitoring}

\shortauthors{abbasi et al.}
\title [mode = title]{An Innovative Data-Driven and Adaptive Reinforcement Learning Approach for Context-Aware Prescriptive Process Monitoring}

\author[1]{Mostafa Abbasi}
\ead{abbasi@uvic.ca}

\author[1]{Maziyar Khadivi}
\ead{mazy1996@uvic.ca}

\author[1]{Maryam Ahang}
\ead{maryamahang@uvic.ca}

\author[2]{Patricia Lasserre}
\ead{patricia.lasserre@ubc.ca}

\author[2]{Yves Lucet}
\ead{yves.lucet@ubc.ca}

\author[1]{Homayoun Najjaran}
\cormark[1]
\ead{najjaran@uvic.ca}

\address[1]{Faculty of Engineering and Computer Science, University of Victoria, BC, Canada}

\address[2]{Computer Science, Irving K. Barber Faculty of Science, University of British Columbia, BC, Canada}

\cortext[cor1]{Corresponding author}

\begin{abstract}
The application of artificial intelligence (AI) and machine learning (ML) in business process management has advanced significantly; however, the full potential of these technologies remains largely unexplored, primarily due to challenges related to data quality and availability. We present a novel five-step framework called Fine-Tuned Offline Reinforcement Learning Augmented Process Sequence Optimization (FORLAPS), which aims to identify optimal execution paths in business processes by leveraging reinforcement learning enhanced with a state-dependent reward shaping mechanism, thereby enabling context-sensitive prescriptions. We implemented this approach on real-life event logs from an energy regulator in Canada and other real-life event logs, demonstrating the feasibility of the proposed method. Additionally, to compare FORLAPS with the existing models (Permutation Feature Importance and multi-task Long Short Term Memory model), we experimented to evaluate its effectiveness in terms of resource savings and process time span reduction. The experimental results on real-life event log validate that FORLAPS achieves 31\% savings in resource time spent and a 23\% reduction in process time span. To further enhance learning, we introduce an innovative process-aware data augmentation technique that selectively increases the average estimated Q-values in sampled batches, enabling automatic fine-tuning of the reinforcement learning model. Robustness was assessed through both prefix-level and trace-level evaluations, using the Damerau–Levenshtein distance as the primary metric. Finally, the model’s adaptability across industries was further validated through diverse case studies, including healthcare treatment pathways, financial services workflows, permit applications from regulatory bodies, and operations management. In each domain, the proposed model demonstrated exceptional performance, outperforming existing state-of-the-art approaches in prescriptive decision-making, demonstrating its capability to prescribe optimal next steps and predict the best next activities within a process trace.
\end{abstract}

\maketitle
% \end{frontmatter}
\section{Introduction}
\label{sec:intro}
Machine learning (ML) has been significantly impacting business processes in recent years, addressing a variety of objectives such as remaining time prediction, next activity prediction, and drift detection \citep{weinzierl2024machine}.  These objectives typically aim to improve one or more of the four key performance dimensions: quality, cost, time, and flexibility. The use of process-aware information systems and rich historical data has led to an increasing trend in implementing ML/AI methods for process improvement projects \citep{abbasi2024review}.
% Businesses constantly strive to optimize costs and time, particularly in their operational processes. Hence, scheduling, sequencing, and queuing activities are pivotal in achieving these goals. The advent of object-centric process mining has further emphasized the importance of these efforts by accurately capturing the complexity of real-world business processes, where multiple activities often occur simultaneously.\cite{vanderAalst2019ObjectCentricPM, khandaker2024transformer}. While there is a substantial body of research on scheduling problems, such as job-shop and flow-shop scheduling, these methods often fall short in practical business settings. Variations in occurrence frequency, shorter durations, and increased uncertainty distinguish business process scheduling from traditional problems. Thus, the unique challenges in business process scheduling demand innovative approaches that account for extensive data and the specific nature of process activities.

The importance of activity sequencing and best next activity prediction is paramount across diverse business domains. In healthcare, the order of diagnostic and treatment activities directly affects patient outcomes, length of stay, and resource utilization \citep{neuberger2024leveragingdataaugmentationprocess}. In financial services, the sequencing of compliance checks influences both regulatory adherence and processing times. Manufacturing environments face similar challenges, where production activity sequencing directly influences throughput, quality metrics, and resource utilization across integrated supply chains. Similarly, manufacturing environments rely on well-ordered production activities to optimize throughput, maintain quality standards, and manage resources efficiently. These cross-domain applications underscore how activity sequencing decisions can significantly impact performance outcomes. This interdependence is particularly evident in object-centric processes where multiple departments interact with shared business objects, creating complex coordination requirements that significantly affect process outcomes. In such highly interdependent contexts, determining the optimal sequence and timing of activities becomes essential to maintain process resilience and performance. Each activity decision not only affects immediate results but also shapes the downstream trajectory of the process.  Therefore, approaches that can identify optimal activity pathways while considering both immediate and downstream effects are increasingly crucial for effective process management \citep{kratsch2021machine}.

Traditional methods for activity and task sequence optimization have employed exact, heuristic, and meta-heuristic techniques over the past decades \citep{Khadivi2024AMM, khadivi2023deepreinforcementlearningmachine}. However, these methods often fall short in capturing the inherent uncertainties within process models.

To address these uncertainties, Predictive Process Monitoring has emerged as a powerful tool for forecasting outcomes at the onset or during process execution \citep{GOOSSENS2024121263}. While predictive methods are effective at identifying patterns in data, they do not directly address inefficiencies or offer actionable improvements \citep{PASQUADIBISCEGLIE2022250}. In response, \textit{Prescriptive Process Monitoring}, a more advanced branch of process mining, goes beyond prediction by analyzing It offers actionable recommendations for task sequences and suggests the best next steps to achieve objectives like reducing cycle time, minimizing defects, or avoiding negative outcomes. For example, \cite{weinzierl_prescriptive_2020} introduced a prescriptive business process monitoring (PrBPM) model that optimizes "next best actions" based on key performance indicators (KPIs), combining LSTM-based prediction with business process simulation to enhance throughput time. Unlike purely predictive models, PrBPM aligns recommendations with business objectives by learning from historical event logs. Other studies have employed techniques such as linear temporal logic for optimal path discovery ~\citep{donadello_outcome_2023}, and resource reallocation strategies in high-risk cases~\citep{shoush2021prescriptive}. These interventions target points in the process where actions can significantly reduce cycle times or improve resource utilization. In another work, the importance of activity location was investigated using Permutation Feature Importance (PFI) as an eXplainable AI (XAI) technique \citep{vazifehdoostirani2023uncovering}. This research assessed the impact of permuting activities within a trace on the process outcome, which was treated as a binary class. However, this work did not consider continuous outcomes nor propose a generalizable optimization framework.

% For instance, \cite{shoush2021prescriptive} combines XGBoost for predicting negative process outcomes with causal inference to trigger selective interventions while accounting for resource constraints. Similarly, \cite{bozorgi2021prescriptive} employs orthogonal random forests to estimate the impact of binary interventions on individual cases, focusing on reducing cycle time. This approach identifies and applies interventions precisely where they are most likely to achieve effective cycle time reduction.

While ML has shown promise in prescriptive analytics, Reinforcement Learning (RL) offers distinct advantages in handling sequential decision-making and adapting to dynamic environments. Its ability to react to process drift makes RL especially suitable for prescriptive process monitoring. In this context, \citet{weytjens2023timing} examined prescriptive process monitoring by comparing RL and causal inference (CI) for process optimization through timed interventions using synthetic data. They implemented RL with a Q-learning model and CI with a multi-layer neural network (NN), ensuring a balanced comparison by using the same architecture for both methods. Similarly, \citet{metzger2020triggering}  used online RL for proactive adaptation, identifying the best moments to trigger changes. Despite these advancements, major challenges remain, such as reliance on fixed prediction models and the high cost of data needed for effective RL convergence.

Online RL requires continuous interaction with the environment to learn policies. However, many domains, such as medical treatment, operations research, and supply chain management, require resource-intensive and time-consuming online interactions. In general, these highly dynamic domains can benefit from an offline setting, where the model is trained using pre-collected data, reducing the need for real-time interactions. Offline Reinforcement Learning (also known as data-driven RL) allows for finding the best policy using a pre-collected dataset \citep{Levine2020offline, YU2024121146}. In the medical treatment process, \cite{verhoef2023using} applied prescriptive process monitoring through RL to optimize staff actions in dynamic care processes, particularly when managing client aggression. However, purely offline approaches may become outdated as processes evolve, necessitating integration with online fine-tuning \citep{DENG2023221}. To handle this, \cite{bozorgi2023learning} implements an online RL approach complemented by a data enhancement method, using a Realcause generative model to produce alternative outcomes for each case prefix, allowing the RL agent to explore different actions offline. They integrate causal inference—using causal forests to estimate treatment effects with confidence intervals—enabling the agent to interact online by applying treatments only in cases likely to benefit. Despite its innovation, this approach suffered from limitations such as generative model inaccuracies and a binary action space (treatment or no treatment), which reduced its generalizability.

In \Cref{relatedworks}, we categorize the current landscape of prescriptive process monitoring into two primary objectives: (1) intervention timing and (2) next-best activity recommendation. While reinforcement learning has shown significant promise in both next-best activity recommendation and intervention timing, existing models typically suffer from several limitations. Most approaches are tailored to categorical outcomes, rely on expert-configured environments, and fall short in optimizing for continuous performance indicators such as resource usage and process duration. Moreover, they often fail to capture the sequential dependencies and cost structures inherent in real-world processes, particularly in cases involving long or resource-intensive traces.

Another critical challenge lies in the limited availability of high-quality, labeled datasets, which restricts the effective training and generalization of RL models. This data scarcity often forces reliance on simplified models or purely historical data, leading to convergence issues and limited robustness.

\begin{table}[b!]
\caption{Related work in prescriptive process monitoring}
\label{relatedworks}
\resizebox{500pt}{!}{%
\begin{tabular}{ccccccccccccc}
\hline
\multirow{3}{*}{Papers}      & \multicolumn{3}{c}{Machine Learning}                                                                                                               & \multirow{3}{*}{\begin{tabular}[c]{@{}c@{}}Causal\\ Inference\\  (CATE)\end{tabular}} & \multicolumn{4}{c}{Reinforcement Learning}                                                                                                                                                                                                                   & \multicolumn{2}{c}{Goal}                                                                                                                                        & \multicolumn{2}{c}{Action}                                                                                                   \\ \cline{2-4} \cline{6-13} 
                             & \multirow{2}{*}{LSTM} & \multirow{2}{*}{XGBoost} & \multirow{2}{*}{\begin{tabular}[c]{@{}c@{}}Decision \\ Tree /\\  Random\\  Forest\end{tabular}} &                                                                                       & \multicolumn{3}{c}{Online RL}                                                                                                                                                                                                  & \multirow{2}{*}{Offline RL} & \multirow{2}{*}{\begin{tabular}[c]{@{}c@{}}Categorical \\ Outcome\end{tabular}} & \multirow{2}{*}{\begin{tabular}[c]{@{}c@{}}Continuous\\ Outcome\end{tabular}} & \multirow{2}{*}{\begin{tabular}[c]{@{}c@{}}Best Next \\ Activity/\\ Activities\end{tabular}} & \multirow{2}{*}{Intervention} \\ \cline{6-8}
                             &                       &                          &                                                                                                 &                                                                                       & \begin{tabular}[c]{@{}c@{}}ML\\  Based \\ Environment\end{tabular} & \begin{tabular}[c]{@{}c@{}}Possibilistic\\ Model\\  Environment\end{tabular} & \begin{tabular}[c]{@{}c@{}}Causal \\ Inference\\  Environment\end{tabular} &                             &                                                                                 &                                                                               &                                                                                              &                               \\ \hline
\citep{weinzierl_prescriptive_2020}            & x                     &                          &                                                                                                 &                                                                                       & x                                                                  &                                                                              &                                                                            &                             & \begin{tabular}[c]{@{}c@{}}Cycle Time\\  Violation\end{tabular}                 &                                                                               & x                                                                                            &                               \\
\citep{donadello_outcome_2023}      &                       &                          & x                                                                                               &                                                                                       &                                                                    &                                                                              &                                                                            &                             & \begin{tabular}[c]{@{}c@{}}Process-Specific\\  Outcome\end{tabular}             &                                                                               & x                                                                                            &                               \\
\citep{vazifehdoostirani2023uncovering}  &                       & x                        &                                                                                                 &                                                                                       &                                                                    &                                                                              &                                                                            &                             & \begin{tabular}[c]{@{}c@{}}Process-Specific\\  Outcome\end{tabular}             &                                                                               & x                                                                                            &                               \\
\citep{metzger2020triggering}         & x                     &                          &                                                                                                 &                                                                                       & x                                                                  &                                                                              &                                                                            &                             & \begin{tabular}[c]{@{}c@{}}Process-Specific\\  Outcome\end{tabular}             &                                                                               &                                                                                              & x                             \\
\citep{agarwal2022goal}          & x                     &                          &                                                                                                 &                                                                                       & x                                                                  &                                                                              &                                                                            &                             & \begin{tabular}[c]{@{}c@{}}Cycle Time\\  Violation\end{tabular}                 &                                                                               & x                                                                                            &                               \\
\citep{shoush2021prescriptive}                  &                       & x                        & x                                                                                               & x                                                                                     &                                                                    &                                                                              &                                                                            &                             & \begin{tabular}[c]{@{}c@{}}Process-Specific\\  Outcome\end{tabular}             &                                                                               &                                                                                              & x                             \\
\citep{bozorgi2021prescriptive}         &                       &                          & x                                                                                               & x                                                                                     &                                                                    &                                                                              &                                                                            &                             & \begin{tabular}[c]{@{}c@{}}Process-Specific\\  Outcome\end{tabular}             &                                                                               &                                                                                              & x                             \\
\citep{bozorgi2023learning}          &                       &                          &                                                                                                 &                                                                                       & x                                                                  &                                                                              & x                                                                          &                             & \begin{tabular}[c]{@{}c@{}}Net gain\\  (Gain - Cost)\end{tabular}               &                                                                               &                                                                                              & x                             \\
Our work                     &                       &                          &                                                                                                 &                                                                                       &                                                                    &                                                                              &                                                                            & x                           & \begin{tabular}[c]{@{}c@{}}Process-Specific\\  Outcome\end{tabular}             & \begin{tabular}[c]{@{}c@{}}Wasted \\ Resource\end{tabular}                    & x                                                                                            &                               \\ \hline
\end{tabular}}
\end{table}

To mitigate data scarcity, researchers have explored reducing model complexity or using simpler algorithms, often at the expense of accuracy or convergence. While model-based RL  holds promise, its black-box nature and dependence on high-dimensional data limit its practical applicability. Conversely, model-free RL is more interpretable and adaptable but relies heavily on reliable interaction data for stable convergence. Some approaches incorporate auxiliary ML models to address data limitations, but these often degrade prediction accuracy or generate poor-quality inputs. Ultimately, such shortcomings hinder the efficient discovery of optimal execution paths and compromise model robustness. 

As outlined in \Cref{relatedworks}, our proposed framework, FORLAPS (Fine-Tuned Offline Reinforcement Learning Augmented Process Sequence Optimization), addresses these challenges by identifying optimal next activities through a five-step framework that combines offline RL with process-aware data augmentation. By alleviating data scarcity and generating a prioritized set of activity sequences, FORLAPS offers greater flexibility and actionable insights for decision-makers.

A critical limitation in existing approaches is their inability to handle varying prefix lengths across traces, which can lead to different process outcomes and carry distinct costs. Prior models often ignore the continuous nature of key performance indicators (KPIs), such as the number of redundant activities or total execution time—both of which significantly affect process optimization and downstream decision-making. To address this gap, our model introduces a state-dependent reward shaping mechanism, configurable as length-aware or duration-aware. This allows the RL agent to impose higher penalties for inefficient behavior in long or costly traces, while de-emphasizing less critical cases. This dynamic reward structure supports context-aware prescriptions, enabling the model to generalize more effectively across heterogeneous business processes.

Moreover, continuous outcome metrics are incorporated into the RL formulation, allowing for more nuanced evaluation and comparison between models. FORLAPS improves exploration, accelerates convergence, and enhances training stability, making it applicable across various domains—such as healthcare, supply chain management, and operations, where decisions must often be made under uncertainty with limited historical data. Our main contributions are listed as follows:
\begin{itemize}

\item We propose a framework that goes beyond traditional binary interventions and static constraints, achieving measurable improvements in both categorical outcomes (\textit{e.g.,} process success/failure) and continuous KPIs (\textit{e.g.,} resource consumption and process duration). Within the RL framework, we design an adaptive, state-dependent reward function that favors efficient execution by penalizing costly or lengthy traces.

\item Novel process-aware data augmentation that generates synthetic process traces while preserving business constraints and process semantics, a significant improvement over methods relying solely on historical data.

\item  Fine-tuned offline reinforcement learning framework that integrates augmented and historical data, enhancing policy learning and exploratory capabilities without requiring extensive online interactions.

\end{itemize}

The remainder of this paper is organized as follows:
Section 2 presents preliminaries and outlines a motivating example derived from our collaboration with an industry partner, followed by a detailed description of our methodology.
Section 3 discusses the experiments conducted on the real-life case study and analyzes the corresponding results.
Section 4 extends the evaluation to public datasets to assess the generalizability of our approach and reports validation outcomes.
Section 5 concludes the paper, summarizing key findings and outlining directions for future research.

\section{Motivation and problem formulation}
\subsection{Motivating example}
The dataset used in this study originates from a Canadian energy regulator responsible for overseeing the approval processes within the oil and gas industry. The data was extracted and transformed into an event log format, capturing all submitted applications since 2016. The resulting log contains 70,164 events across approximately 14,000 cases, each representing a distinct application. Each application follows a complex, partially parallel process involving up to 12 distinct review activities, which may be executed concurrently. The specific combination of required activities varies by case, introducing significant process heterogeneity. A key challenge arises from the interdependencies between activities: if any activity within a case is disapproved, it may trigger the disapproval of other related activities,  whether still in progress or previously approved, resulting in costly inefficiencies. These cascading failures often result in inefficiencies, resubmissions, and prolonged approval timelines. Some cases have undergone up to 10 revisions, highlighting the urgent need to optimize activity sequencing to minimize rework and improve overall process efficiency. Furthermore, the complexity of this process is reflected in the number of potential activity sequences. Theoretically,  there are $2^{12} - 12$ possible activity combinations, yet only 790 unique activity sequences have been observed over the years. This bounded but diverse action space makes it feasible to adopt a model-free reinforcement learning approach to estimate Q-values for identifying optimal activity sequences. Without such optimization, inefficiencies persist, delaying critical approvals and increasing operational costs for both the regulator and applicants. By learning from historical execution patterns and outcomes, a well-trained RL model can recommend more efficient activity orders, reducing the risk of rejection cascades and improving overall process performance.

% \Cref{tab:sampletrace} presents representative examples of training event log traces extracted from the case study.

% \begin{table}[h]
%     \centering
%     \small
%     \caption{Some sample training traces in the event log}
%     \label{tab:sampletrace}
%     \begin{tabular}{p{2.2cm} p{7.8cm}}
%         \toprule
%         \textbf{ID} & \textbf{Trace} \\ \midrule
%         1 & (ARR, FR, CNR, DPER, FER, FNC, AGR, LHR, CRR) \\
%         2 & (CRR, FER) \\
%         3 & (FNC, CNR, DPER) \\
%         \bottomrule
%     \end{tabular}
% \end{table}

\subsection{Data-driven (Offline) Reinforcement Learning}

Reinforcement learning requires interaction between the agent and the environment. Consequently, the Markov Decision Process (MDP) provides a framework to describe and formulate the environment.

{\textbf{Definition:}} In a MDP formulation, a tuple $\mathcal{M} = (\mathcal{S}, \mathcal{A}, T, d_0, r, \gamma)$ is defined, where the system comprises distinct states ($\mathcal{S}$) and actions ($\mathcal{A}$). Each state represents a specific configuration of activities. To guide the process toward optimal completion, the transition function $(T)$ specifies the probability of moving from state $S_t$ to $S_{t+1}$ given an action ($a_t$). $d_0$ denotes the initial state and $\gamma$ represents the discount factor.

As previously discussed, each episode, representing instances of the process, presents a dynamic set of available activities for initiation (\textit{e.g.,} $\sigma = <(FN, LHR, ARR, AGR), Revision 0>$).

As \Cref{psuedocode} shows,  the process transitions to a subsequent state $s'$ based on the outcome of the executed task.

If the selected activity is approved, a positive reward ($r$) is assigned, and the task is removed from the list of remaining activities. If the activity is disapproved, the agent receives a penalized reward, calculated as $-r$ multiplied by the number of previously executed activities. This reward structure encourages early identification of faulty or high-risk paths to minimize wasted effort. By penalizing disapprovals that occur later in the trace, the model learns to front-load risky decisions, identifying critical bottlenecks earlier and reducing the risk of cascading failures. At this stage, using \Cref{qupdate}, we update the value of $Q(s, a)$ using 
\begin{equation}
\label{qupdate}
Q(s, a) \gets Q(s, a) + \alpha \cdot \left( reward + \gamma \cdot \max_{a'} Q(s', a') - Q(s, a) \right).
\end{equation}
To monitor the performance of the model, we track the moving average of the trained Q-values over the time steps.

We aim to determine the optimal policy or probability distribution for a particular action given specific states or conditions, denoted as $(\pi(a_t|s_t))$, through the interaction of an agent with the MDP $\mathcal{M}$. This involves employing a behavior policy based on the dataset. We use ($\pi_\beta$) to denote the distribution over states and actions in $D$, where we assume that the state-action tuples (s, a) $\in D$.

In our case study, the reward function is designed to discourage late-stage failures by assigning increasingly negative rewards to disapproved activities that occur deeper in the execution trace. This reflects the principle that the impact of failure increases with process depth, especially when significant resources have already been committed. The agent is thus incentivized to uncover risks early, aligning with the fast-fail principle commonly adopted in industrial decision-making to reduce cost and time overhead. In \Cref{psuedocode}, the MDP formulation is outlined.

Let $S$ represent the set of all possible states and $A$ represent the set of all possible actions. In this setup, the Q-learning process updates the Q-value $Q(s, a)$ for each state-action pair $(s, a)$, using the Bellman equation \Cref{qupdate}.

A state $s_t$ at time step $t$ is represented as a sequence of actions: $s_t = (a_0, a_1, \ldots, a_t).$ To further guide learning, we introduce a state-dependent reward shaping mechanism, also referred to as a length-aware reward function. Unlike static punishments, this method applies progressive penalties as the trace length increases, encouraging the agent to behave more cautiously in longer or risk-prone sequences. This dynamic shaping allows the model to differentiate between early and late failures, reinforcing the value of front-loading uncertainty and enabling context-aware prescriptions.

Training progress is evaluated using a sliding window of size $w$, which captures Q-value trends over recent time steps, providing insights into convergence and policy stability.

We define a state-dependent reward shaping function as:
\begin{equation}
R(s_t, a_t) = -r \cdot |s_t| \cdot \mathcal{K}(s_t, a_t)
\end{equation}
where $\mathcal{K}(s_t, a_t)$ is an indicator function that returns 1 if the action $a_t$ leads to a KPI violation, and 0 otherwise.
Here, $|s_t|$ is the number of activities executed up to state $s_t$, and $r > 0$ is a constant base penalty. The reward grows more negative as the process advances and a KPI violation happens, hence providing dynamic shaping depending on the trace length.

\begin{algorithm}
\begin{algorithmic}[1]
\caption{Q-Learning for Process Trace Sequence}
\label{psuedocode}
\State \textbf{Parameters:} $\alpha$, $\gamma $
\State Initialize $Q(s, a)$ for all $s \in S$ (possible states) and $a \in A(s)$ as an empty list
\State $\forall \sigma$ Initialize remaining activities in $\sigma$ as $RT$
    \While {$RT$}
        \State Choose $\mathcal{A}$ using $\pi_\beta$
        \If {$User Task Status \Rightarrow \text{Approved}$}
            \State $reward \gets r  $
            \State $RT \gets RT \setminus \{\mathcal{A}\}$
        \ElsIf {$User Task Status \Rightarrow \text{Disapproved}$}
            \State $reward \gets -r \cdot |s_t| \cdot \mathcal{K}(s_t, a_t)$
        \EndIf
        \State $Q(s, a) \gets Q(s, a) + \alpha \cdot \left( reward + \gamma \cdot \max_{a'} Q(s', a') - Q(s, a) \right)$
    \EndWhile
\end{algorithmic}
\end{algorithm}

\subsection{Process data augmentation}

To enhance the performance of the offline reinforcement learning (RL) algorithm, we introduce a novel process-aware data augmentation technique that systematically generates synthetic process traces while preserving business constraints and process semantics. Unlike traditional augmentation methods, which often focus solely on introducing noise or random variations, our method updates standard data augmentation by considering the logical dependencies, precedence, and conflicts between activities. This ensures that generated synthetic traces are not only diverse but also realistic and aligned with real-world process behaviors.

The augmentation technique enriches the dataset (event logs $\mathcal{L}$) by incorporating variability through the addition of random noise ($\varepsilon$) to timestamp fields, altering them by up to 10\% of the time differences, which affects the order of activities. In other words, let \( \Delta t_{ij} \) represents the time difference between two consecutive activities \( A_i \) and \( A_j \) in a given trace. The timestamp for activity \( A_j \) can be modified as:
\begin{equation}
{
{Timestamp}_{new}(A_j) = Timestamp(A_j) + \varepsilon \times \Delta t_{ij}
}
\end{equation}

where \( \varepsilon \) is a random noise factor uniformly sampled from the interval \([-0.1, 0.1]\) to adjust the timestamp within a 10\% range. This modification induces variability while respecting the temporal ordering of activities.

Additionally, we focus on valuable data by randomly removing 5\% of the groups where all activities are completed and selectively removing certain activities within the groups. Within these selected groups, we selectively remove non-critical activities, excluding those with fewer than three activities or known to be essential to process execution. This preserves core process behavior while introducing additional variation for training.

To ensure compliance with business rules and process semantics, we implement the following filtering mechanism that respects activity dependencies, precedence, and conflicts. We define the business rules \( \mathcal{B} \) as a set of logical constraints, which include: \textbf{Activity Precedence}: If activity \( \mathcal{A}_i \) must precede \( \mathcal{A}_j \), this relationship must be preserved during augmentation.  \textbf{Activity Conflict}: If activities \( \mathcal{A}_i \) and \( \mathcal{A}_j \) conflict and should not appear together in the same trace, they must not be co-occurring. \textbf{Activity Co-occurrence}: If activities \( \mathcal{A}_i \) and \( \mathcal{A}_j \) depend on each other for existence, they must appear together in the same trace.

Let the predefined business constraints be represented as a set \( \mathcal{B} \), such that: $\mathcal{B} = \{b_1, b_2, \dots, b_k\}$. Where each \( b_i \) represents a business rule or process constraint. For a given process trace \( \sigma \), we define the semantic validity ($\mathcal{SV}$) of ( $\sigma$ ) as:

\begin{equation}
\forall \sigma' \subseteq \mathcal{L}',\ \forall b \in \mathcal{B}:\ \mathcal{SV}(\sigma', b) 
\quad \text{and} \quad 
\exists \sigma' \subseteq  \mathcal{L}',\ \exists b \in \mathcal{B} \text{ such that } \sigma' \nsubseteq b \Rightarrow  \mathcal{L}' \setminus \{\sigma'\}
\label{semantic}
\end{equation}

Where $\sigma'$ denotes a trace in the augmented log $\mathcal{L}'$. \Cref{semantic} ensures that the augmented data remains within the bounds of process semantics and business constraints.

Finally, we address the potential overestimation of Q-values for out-of-distribution (OOD) actions by ensuring that the augmented dataset not only provides diversity but also preserves the structural and functional validity of the process flows. By incorporating business rules and process semantics into the augmentation process, we mitigate the risk of introducing unrealistic traces into the dataset, maintaining the alignment between synthetic and real-world process behavior \citep{yang2022rorl}.

\begin{figure}[h]
    \centering
    \includegraphics[width=1\linewidth]{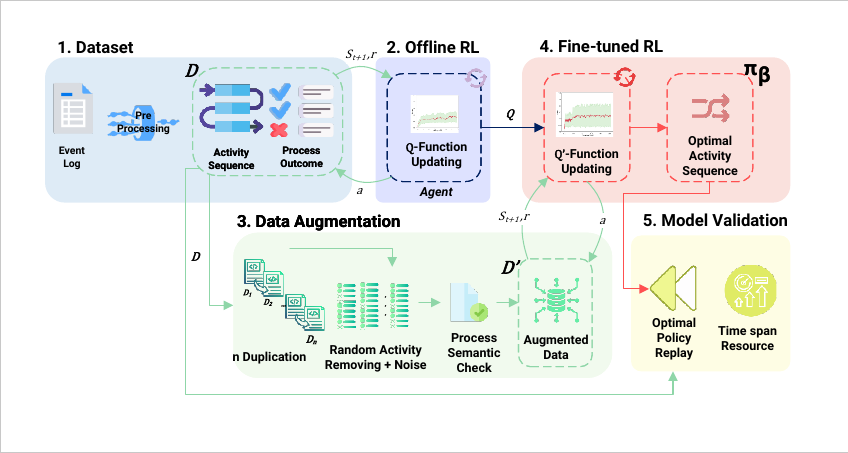}
    \caption{FORLAPS framework~\scriptsize{(Icons from www.flaticon.com)}}
    \label{framework}
\end{figure}

Beyond initial offline training, we enhance policy learning through a fine-tuning phase conducted entirely in the offline setting. Rather than relying on online environment interactions, we simulate exploration by incorporating synthetically augmented traces into the training process. This strategy addresses common challenges in offline RL, such as limited state-action coverage and distributional shift, by diversifying the dataset in a process-aware manner. After estimating Q-values from the original event logs, the model is fine-tuned using the augmented data, enabling iterative updates to the Q-function ($Q'(s, a)$). To evaluate the impact of this augmentation-driven fine-tuning, we conducted experiments using datasets expanded with different augmentation scales, comparing Q-value convergence and policy performance across isolated and combined data scenarios. This refinement process facilitates the discovery of an improved policy (\Cref{optimalpi}) while remaining fully compliant with an offline RL paradigm.

\begin{equation}
\label{optimalpi}
\pi^*_{\beta} = argmax_{a} Q'^{\pi}(s, a) = argmax_{\pi} \mathbb{E}_{\pi} \left[ \sum_{t=0}^{K} \gamma^t r(s_t) \right]
\end{equation}
\Cref{framework} shows the 5-step framework that summarizes the process of pre-processing event logs, offline RL training with Q-function updates, data augmentation through duplication, noise addition, and compliance check, fine-tuned RL for optimal policy generation, and model validation for evaluating resource efficiency.
Technical details of this framework and the source code are available on GitHub \footnote{https://github.com/mstabc/FORLAPS}.

To evaluate the efficacy of the FORLAPS framework, we conducted comprehensive experiments using a diverse corpus of business process data. Our experimental methodology leveraged both a primary case study from an energy regulatory commission in Canada and nine publicly available event logs from various domains to ensure external validity and generalizability. The case study dataset includes rich temporal information, activity approval statuses, and resource allocation records, enabling fine-grained analysis of performance across both categorical and continuous process indicators. We structured our experimental evaluation into two complementary phases: first, a comparative performance assessment against established approaches (Permutation Feature Importance and multi-task LSTM models), measuring resource time savings and process time span reduction; second, We introduced a novel validation method using Damerau–Levenshtein distance to assess the similarity between recommended and actual activity sequences. This distance metric captures structural deviations and offers a process-aware measure of model robustness in optimal sequence generation. This dual evaluation strategy demonstrates that FORLAPS not only improves operational efficiency but also consistently recommends high-quality, semantically valid sequences across a variety of process settings. The results provide compelling evidence of its superior performance in prescriptive process monitoring applications across domains such as healthcare, finance, public administration, and industrial operations.

\section{Case study experiments}

\begin{figure}[htbp]
    \centering
        \begin{subfigure}[b]{0.24\linewidth}
        \centering
        \includegraphics[width=\linewidth]{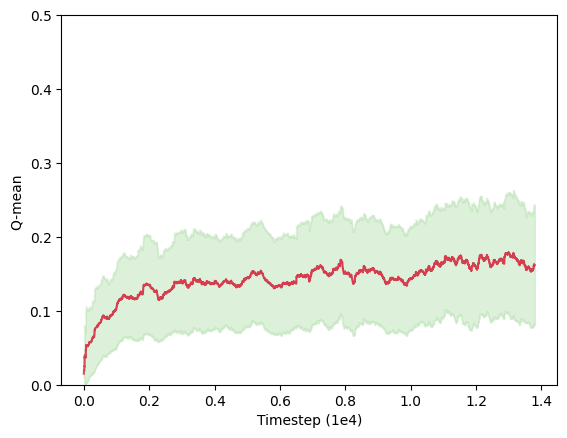}
        \caption{Offline RL}
        \label{fig:bcerl}
    \end{subfigure}
    \begin{subfigure}[b]{0.24\linewidth}
        \centering
        \includegraphics[width=\linewidth]{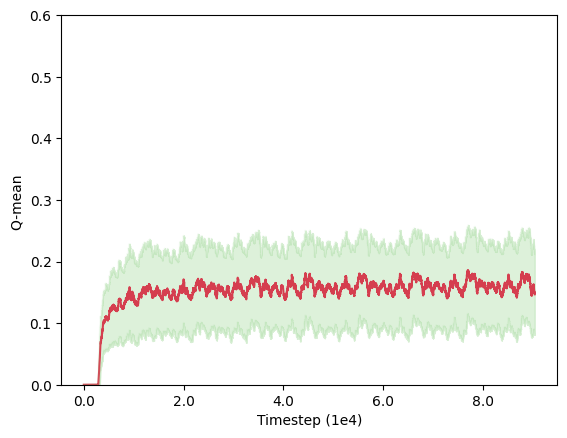}
        \caption{Isolate 100K}
        \label{fig:100k}
    \end{subfigure}
    \begin{subfigure}[b]{0.24\linewidth}
        \centering
        \includegraphics[width=\linewidth]{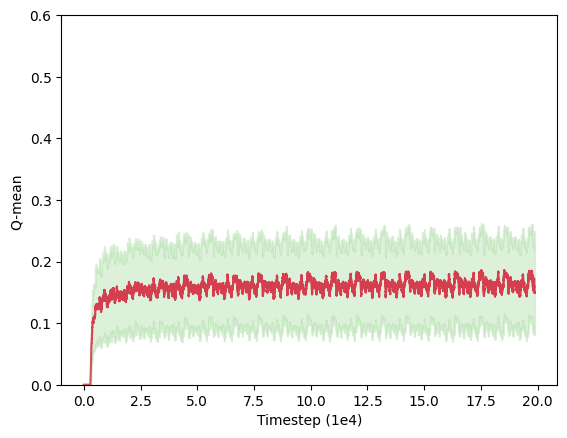}
        \caption{Isolate 200K}
        \label{fig:200k}
    \end{subfigure}
    \begin{subfigure}[b]{0.24\linewidth}
        \centering
        \includegraphics[width=\linewidth]{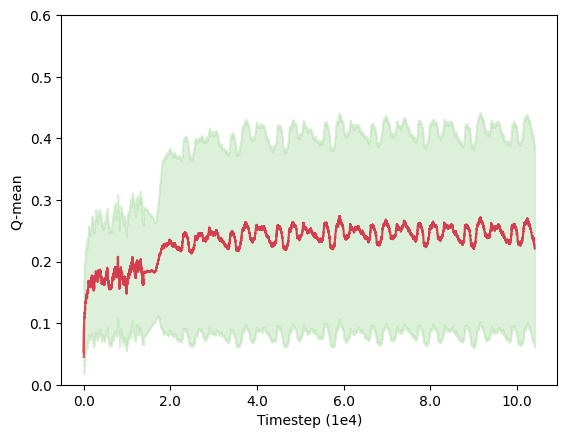}
        \caption{Fined-Tuned RL}
        \label{fig:offonfinetuning}
    \end{subfigure}

    \caption{Mean Q-values ($\pm$~std) across timesteps }
    \label{fig:comparison}
\end{figure}

We evaluate our algorithm's performance on the case study by tracking the evolution of Q-values over successive episodes using static datasets. As shown in \Cref{fig:bcerl}, the average Q-value progressively increases throughout the training episodes, with the shaded regions representing the standard deviation. This upward trend indicates that the model effectively learns an improved policy over time. The observed improvement in policy quality corresponds to higher estimated Q-values, providing empirical evidence that the model's value estimation reliably converges as training proceeds.

We conducted a comprehensive hyperparameter search to analyze the impact of key parameters such as learning rate ($\alpha \in [0,1)$) and discount factor ($\gamma \in [0,1)$) on the model's performance and set the optimal values.

In the next phase of our study, the goal is to enable the agent to achieve higher average Q-value estimates by leveraging offline pre-training, followed by further fine-tuning using augmented data. This progressive learning process aims to improve the learned policy through enhanced value estimation within the offline setting.
We evaluated the impact of utilizing behavior policy data both in isolation (\Cref{fig:bcerl}) and in combination with offline fine-tuning, starting from pre-trained Q-values. Initially, the model was trained using static offline data. To enhance generalization and policy quality, we augmented the dataset with an additional 100K and 200K timesteps. As shown in \Cref{fig:offonfinetuning}, this fine-tuning procedure consistently outperforms isolated models for both augmentation scales, converging more rapidly during early training and achieving superior final performance (\Cref{fig:100k,fig:200k}).
Moreover, \Cref{fig:offonfinetuning} confirms that fine-tuning not only preserves the performance of the pre-trained model but also enhances it over time. Importantly, \Cref{fig:comparison} highlights the pivotal role of pre-training: models initialized with pre-trained Q-values significantly outperform purely offline or isolated models even after 100K and 200K steps, demonstrating the effectiveness of representation transfer. Overall, the fine-tuning process yields a 44\% improvement in performance, stabilizing Q-values and reinforcing the benefits of a hybrid learning strategy within a fully offline RL setting.

A key takeaway from our experiments is that even isolated training with 100K and 200K timesteps demonstrates that a relatively small dataset is sufficient to effectively train the agent and uncover meaningful process patterns. This highlights the viability of lightweight training scenarios in data-scarce environments. However, as shown in subsequent validation, this effectiveness is contingent on achieving a balance between the number of unique states and the number of existing traces. When this balance is disrupted—such as in datasets with high state complexity but limited trace diversity—model performance and convergence may degrade due to insufficient exposure to representative state-action transitions.

More broadly, accelerating policy learning through offline data is crucial for real-world operations and business process management, where interaction costs are high and simulations often fail to capture all real-world uncertainties. Our approach leverages a data-enhancement-based fine-tuning strategy, where pre-trained policies are refined using augmented traces rather than online interactions. This enables policy improvement while remaining within a fully offline RL setting.

Additionally, the introduction of controlled noise and augmented process traces supports better generalization and reduces the risk of converging to suboptimal policies. For this reason, we selected a broad range of episodes to train the RL agent more effectively, ensuring robustness in learning sequential strategies across complex business processes.

Finally, after updating the Q-values for each state-action pair, the trained agent is able to prioritize activities based on their estimated utility. \Cref{tab:freq} presents the most frequently encountered states along with the corresponding optimal actions as determined by the learned policy. This policy allows the agent to reorder activities within a case, aiming to minimize inefficiencies by selecting actions that yield higher expected rewards.

Moreover, by considering the experiences and Q-values stored in the buffer, the agent can provide not only a single best next activity, but also a ranked set of top actions, offering flexibility for decision-makers to select from a priority list based on contextual or operational preferences. This allows the system to act as a decision support tool, rather than prescribing rigid sequences.

It is important to note, however, that this sequential reordering may introduce additional delays, as activities are no longer executed in parallel. The implications of this trade-off between efficiency and timeliness will be explored in greater detail in the following section.

\begin{table}[ht]
\caption{Most frequent activity combination and the best activity to initiate}
\label{tab:freq}
\centering
\scriptsize
\begin{tabular}{lcccl}
\cline{1-4}
\textbf{State}                     & \textbf{StateLength} & \textbf{Action} & \textbf{Q} &  \\ \cline{1-4}
(PER, AGR, ARR, FR, LHR, CNR, FNC) & 7                    & FNC             & 0.499691   &  \\
(PER, ARR, FR, LHR, CNR, FNC)      & 6                    & LHR             & 0.499364   &  \\
(ARR, FR, LHR, CNR, FNC)           & 5                    & FR              & 0.495598    \\ \cline{1-4}
\end{tabular}
\end{table}

To assess the effectiveness of the proposed approach, we replayed 15,000 traces from the test set of event logs using the learned optimal policy. Each experience is represented as a transition tuple $(s_t, a_t, r_t, s_{t+1})^j$ for an episode $j \in [1, K]$ at timestamp $t \in [\tau^j, \tau^j + d^j]$, where $d^j$ denotes the duration of the case.

To assess the performance improvements achieved through our RL-based policy, we use two key performance indicators (KPIs) to compare the optimized traces against the original process executions:

\textbf{Saved time span ($d^j$):} This KPI measures the difference between the actual duration of a case and the duration of the case when activities are reordered according to the RL optimal policy.

\textbf{Saved resources time spent $(\Sigma_j \Sigma_t \, d^j)$:} This quantifies the overall reduction in resource engagement time. It compares the cumulative time spent on activities in the original traces versus the time required under the optimized policy, reflecting improvements in resource utilization efficiency.

To evaluate methods for recommending the best next activity, we implemented a Multi-task LSTM-based model designed to optimize key performance indicators (KPIs). Building on the approach proposed by \citet{weinzierl_prescriptive_2020}, this technique consists of an offline training phase involving two components: a Multi-task Process Prediction (MPP) model, which forecasts both the next activity and its associated KPIs using a multi-task deep neural network (DNN); and a Multi-case Similarity (MCS) model, which employs a nearest-neighbor algorithm to retrieve historically similar action sequences.

\begin{figure}[h!]
    \centering
    \begin{subfigure}[b]{0.48\linewidth}
        \centering
        \includegraphics[width=\linewidth]{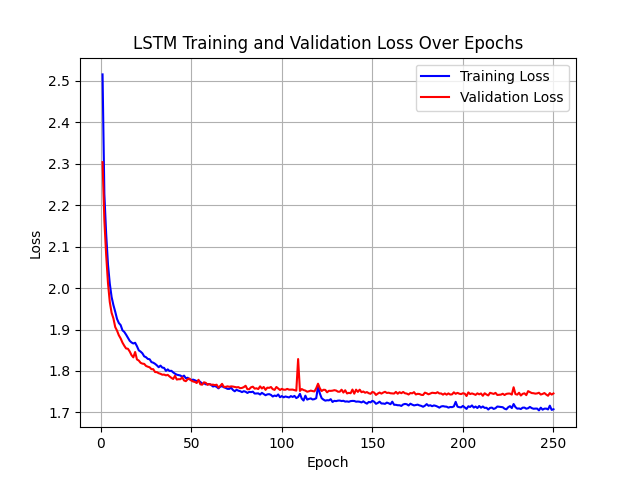}
        \caption{Training and validation loss over 250 epochs.}
        \label{fig:bcerlstm}
    \end{subfigure}
    \begin{subfigure}[b]{0.48\linewidth}
        \centering
        \includegraphics[width=\linewidth]{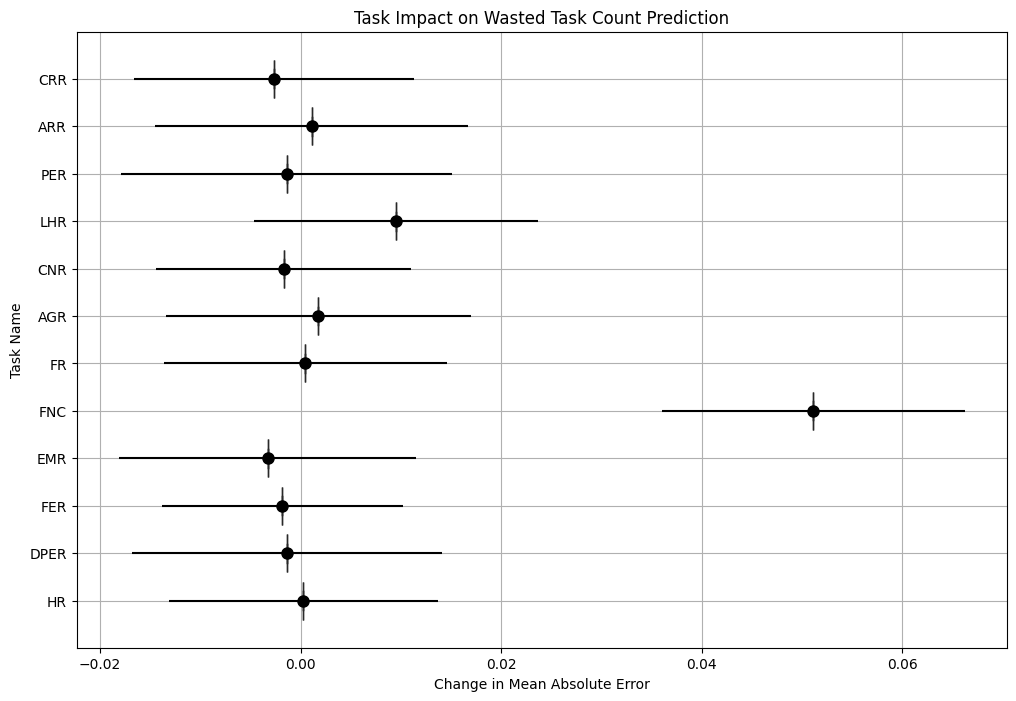}
        \caption{Change in (MAE) $\pm~std$.}
        \label{fig:maechange}
    \end{subfigure}
    
    \caption{LSTM and PFI training results}
    \label{fig:combined}
\end{figure}

In the online phase, these models operate jointly to recommend the most suitable next actions in real time, using patterns extracted from historical event logs to support prescriptive process optimization. As illustrated in \cref{fig:bcerlstm}, the Multi-task LSTM model achieves stable convergence. Model configurations—available in the accompanying GitHub repository—were fine-tuned to prevent overfitting and ensure generalizable performance across diverse process scenarios.

Another model to compare our model's results with the existing approach, we used Permutation Feature Importance (PFI), a method in explainable AI (XAI), to identify the significance of activity locations. In this context, we observe the location of the studied itemset. When permuting itemsets with multiple activities, a random shuffle may change the relative order of activities in the itemset, providing insight into the importance of each activity's location in the trace. Building on the implementation in the previous work \citep{vazifehdoostirani2023uncovering}, we trained an XGBoost model on the event log to assess the impact of permuting activity locations on the mean absolute error (MAE). This allows us to identify which activities contribute most to the number of wasted activities, enabling us to prioritize them accordingly. As illustrated in \Cref{fig:maechange}, changes in MAE reveal how sensitive predictive performance is to the location of specific activities. PFI proves to be an effective, context-aware method for analyzing real-life process data, as it adapts to the structure of current traces. However, its applicability is limited to processes with parallel or loosely coupled activities and does not generalize well to strictly sequential workflows commonly found in real-world scenarios.

To quantify the impact of this next-best activity prediction on resources and application time span, we compared the effectiveness of our method against the LSTM-based model and PFI method. The comparison results are presented in \Cref{vs}. This analysis highlights the advantages of our approach in terms of resource efficiency and reduced application time span. On average, across all evaluated traces, for an 8-day work task, FORLAPS achieved a 31\% improvement in resource time savings for an 8-day work task, while the LSTM model and PFI model achieved 21.5\% and 18\% respectively. All models, however, managed to reduce the overall duration by approximately 24\%.

\begin{table}
\centering
\caption{Comparison of the proposed FORLAPS model with the LSTM-based model and PFI approach in terms of evaluation metrics (Average days ± 96\% confidence interval)}
\label{vs}
\scriptsize
\begin{tabular}{cclclclc}
\hline
\textbf{Models}   & \textbf{Saved resource time spent} &  & \textbf{Opt\%} &  & \textbf{Saved time span} &  & \textbf{Opt\%} \\ \cline{1-2} \cline{4-4} \cline{6-6} \cline{8-8} 
\textbf{FORLAPS} & 2.66 ± 0.06                               &  & 31\%           &  & 9.69 ± 0.09                     &  & 23\%           \\
 \textbf{LSTM}& 1.84 ± 0.01 & & 21.5\%& & 10.1 ± 0.12& &25\%\\
\textbf{PFI}      & 1.58 ± 0.02                              &  & 18\%           &  & 10.02 ± 0.14                   &  & 24\%           \\
\textbf{Baseline} & \multicolumn{3}{c}{8.52 per activity}                  &  & \multicolumn{3}{c}{40.56 per case}           \\ \hline
\end{tabular}
\end{table}
Although the initial assumption was that enforcing a sequential execution of activities would lead to longer case durations, our results reveal that resource allocation efficiency, achieved through optimal activity prioritization across all test set traces, can still yield overall time span savings. This indicates that, despite the sequential arrangement, intelligently ordering tasks enables better resource distribution, thereby offsetting potential delays and improving total process performance.
% \begin{figure}
%     \centering
%     \includegraphics[width=0.75\linewidth]{savetimespanimproved.jpg}
%     \caption{Saved  resource and time span  under optimal policy vs actual dataset }
%     \label{fig:save}
% \end{figure}

\section{Evaluation}

In addition to our case study, we applied the proposed approach to multiple publicly available datasets and benchmarked its performance against established techniques from the literature. This evaluation provides a robust basis for generalizing the problem formulation across diverse scenarios and offers a meaningful comparison within the domain of prescriptive process monitoring.

\subsection{Experimental setup}
In \Cref{datasets}, we defined and discussed different types of outcomes and presented the results for all the datasets. To ensure consistency and comparability, the definition of desired outcomes for each dataset follows those established in prior works referenced accordingly.

\begin{table}[h!]
\caption{Overview of datasets used in the experiments.}
\label{datasets}
\resizebox{480pt}{!}{%
\begin{tabular}{lllllll}
\cline{1-1} \cline{3-3} \cline{5-5} \cline{7-7}
\multicolumn{1}{c}{\textbf{Dataset}}          &  & \multicolumn{1}{c}{\textbf{Description}}                                                                                                     &  & \multicolumn{1}{c}{\textbf{Outcome (Categorical)}}                                                                             &  & \multicolumn{1}{c}{\textbf{Outcome (Continuous)}}                                                                   \\ \cline{1-1} \cline{3-3} \cline{5-5} \cline{7-7} 
Sepsis \citep{vazifehdoostirani2023uncovering} &  & \begin{tabular}[c]{@{}l@{}}Event logs from patients’ registration in the\\ Emergency Room to their discharge from the hospital.\end{tabular} &  & \begin{tabular}[c]{@{}l@{}}Whether the patient returns to the \\ Emergency Room within 28 days of discharge.\end{tabular}      &  & \begin{tabular}[c]{@{}l@{}}Number of wasted activities if the patient \\ returns to the Emergency Room.\end{tabular} \\ \cline{1-1} \cline{3-3} \cline{5-5} \cline{7-7} 
BPIC 2015 \citep{donadello_outcome_2023}       &  & \begin{tabular}[c]{@{}l@{}}5 sets of event logs of building permits\\  processed in five Dutch municipalities.\end{tabular}                  &  & \begin{tabular}[c]{@{}l@{}}Whether the request leads to\\ unsuccessful completion.\end{tabular}                                &  & \begin{tabular}[c]{@{}l@{}}Number of wasted activities resulting\\ from incomplete applications.\end{tabular}       \\ \cline{1-1} \cline{3-3} \cline{5-5} \cline{7-7} 
BPIC 2017                                     &  & \begin{tabular}[c]{@{}l@{}}Event logs of a loan application\\ process in a Dutch financial institution.\end{tabular}                         &  & \begin{tabular}[c]{@{}l@{}}Whether the application is \\ rejected, or canceled.\end{tabular}                                   &  & \begin{tabular}[c]{@{}l@{}}Number of wasted activities\\ due to the rejection of the application.\end{tabular}      \\ \cline{1-1} \cline{3-3} \cline{5-5} \cline{7-7} 
Traffic Fines \citep{donadello_outcome_2023}   &  & \begin{tabular}[c]{@{}l@{}}Event logs of traffic fines processing \\ by the Italian local police force.\end{tabular}                         &  & \begin{tabular}[c]{@{}l@{}}Whether the fine leads to \\ full or partial payment.\end{tabular}                                  &  & \begin{tabular}[c]{@{}l@{}}Number of wasted activities due to \\ unpaid fines.\end{tabular}                         \\ \cline{1-1} \cline{3-3} \cline{5-5} \cline{7-7} 
BPIC 2019 \citep{weinzierl_prescriptive_2020}  &  & \begin{tabular}[c]{@{}l@{}}Event logs of purchase order handling processes in \\ a painting and coating company.\end{tabular}                &  & \begin{tabular}[c]{@{}l@{}}Compliance with the KPI for throughput time\\ (average: 71.52 days; violation or not).\end{tabular} &  & \begin{tabular}[c]{@{}l@{}}Amount of time wasted due to \\ compliance violations.\end{tabular}                      \\ \cline{1-1} \cline{3-3} \cline{5-5} \cline{7-7} 
\end{tabular}}
\end{table}
% As noted earlier, a system must define the desired outcomes, which can be either categorical or continuous. To ensure comparability, we adopted a similar outcome structure (reward setting) as in previous works, such as \cite{donadello_outcome_2023}, aligning the conditions for the comparison. The formulation of the Markov Decision Process (MDP) in each case depends heavily on the properties and characteristics of the datasets.

% To evaluate different methods, we split the dataset into training and test sets, implementing various algorithms to determine which one yields the best performance after applying the best next activity policy in different instances. It is important to note that time compliance was only considered for cases with harmonious types of traces. If the processes are different, applying that compliance across all cases would not be appropriate.
Our evaluation considers two distinct problem settings: episodic and non-episodic. In episodic problems, the process terminates upon reaching a predefined goal or condition, and rewards are typically assigned only at the conclusion of the episode. In contrast, non-episodic problems allow continuous transitions between states, enabling rewards and outcomes to be observed throughout the progression of the episode. This distinction is critical for modeling different process structures and assessing the adaptability of the proposed approach across varying process dynamics.

\subsection{Results and discussion}
This section presents the training and evaluation results for all models, including the LSTM-based approach, offline reinforcement learning, and the proposed FORLAPS framework. For RL-based methods, we report the Q-value progression over training epochs, while for the LSTM model, we report the training and validation loss curves. All models were trained on 80'\% of the dataset, with the remaining 20\% reserved for evaluation, ensuring an unbiased assessment on unseen test data and allowing for a more realistic estimate of generalization performance.

\begin{figure}[h!]
    \centering
    
    \begin{subfigure}[b]{0.3\textwidth}
        \centering
        \includegraphics[width=\textwidth]{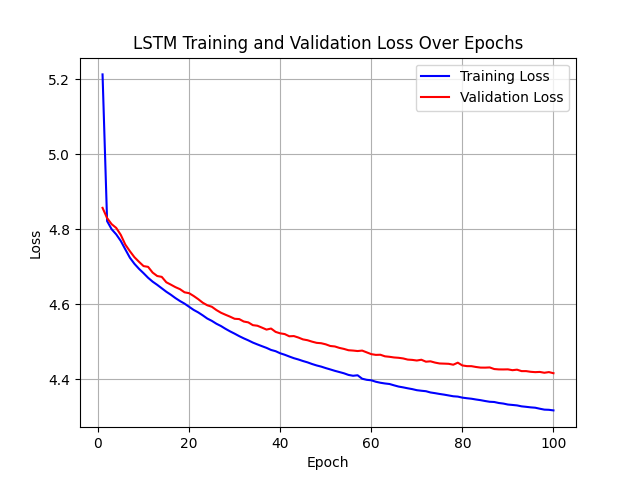}
        \caption{BPIC\_2015\_1}
    \end{subfigure}
    \begin{subfigure}[b]{0.3\textwidth}
        \centering
        \includegraphics[width=\textwidth]{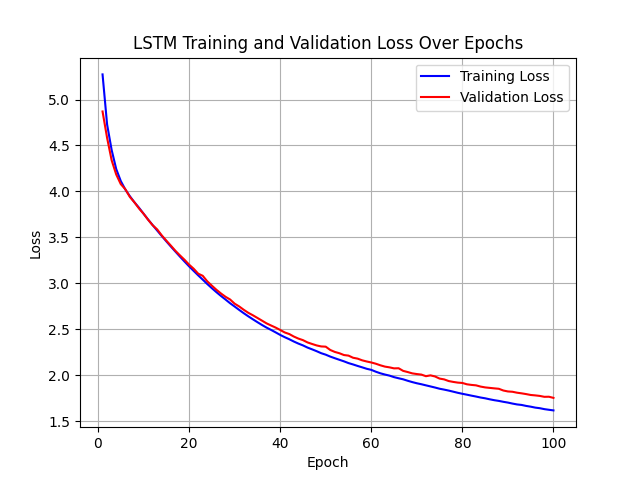}
        \caption{BPIC\_2015\_2}
    \end{subfigure}
    \begin{subfigure}[b]{0.3\textwidth}
        \centering
        \includegraphics[width=\textwidth]{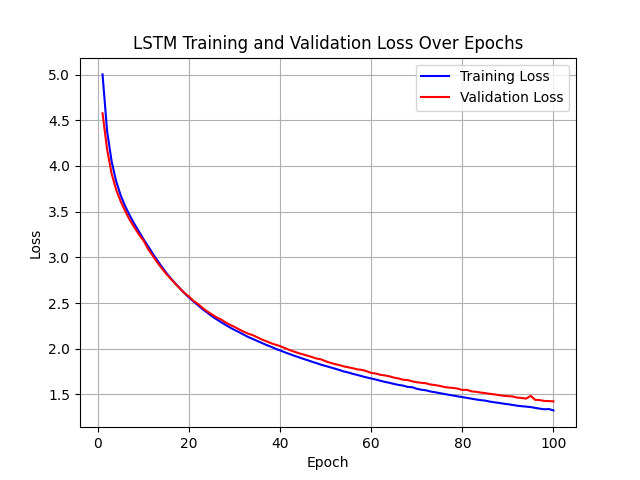}
        \caption{BPIC\_2015\_3}
    \end{subfigure}
    \begin{subfigure}[b]{0.3\textwidth}
        \centering
        \includegraphics[width=\textwidth]{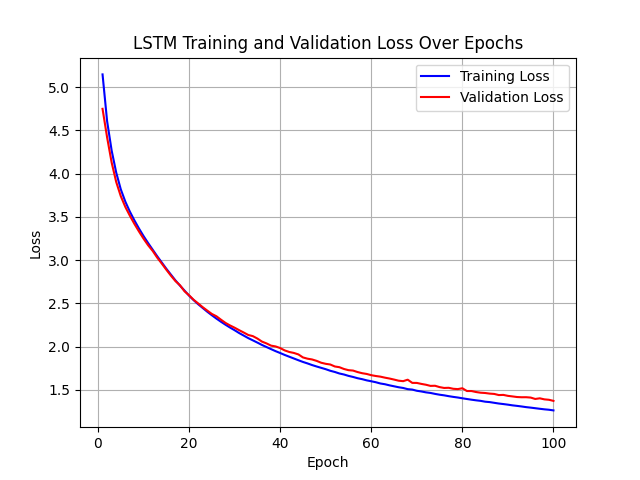}
        \caption{BPIC\_2015\_4}
    \end{subfigure}
    \begin{subfigure}[b]{0.3\textwidth}
        \centering
        \includegraphics[width=\textwidth]{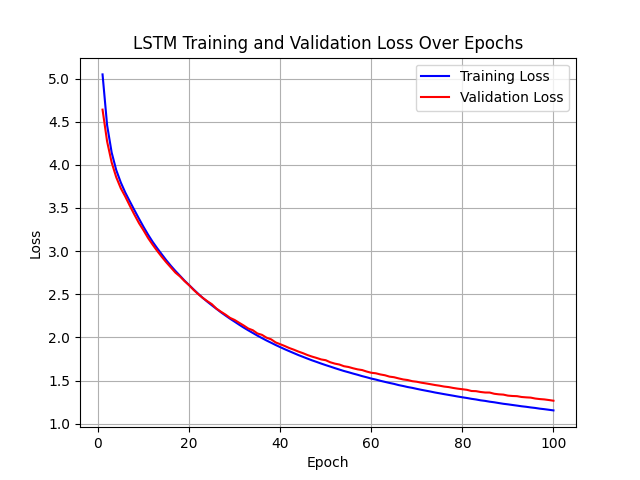}
        \caption{BPIC\_2015\_5}
    \end{subfigure}
    \begin{subfigure}[b]{0.3\textwidth}
        \centering
        \includegraphics[width=\textwidth]{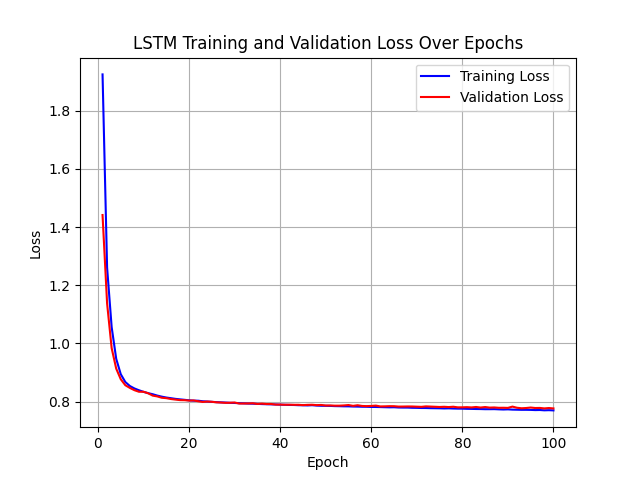}
        \caption{BPIC\_2017}
    \end{subfigure}
    \begin{subfigure}[b]{0.3\textwidth}
        \centering
        \includegraphics[width=\textwidth]{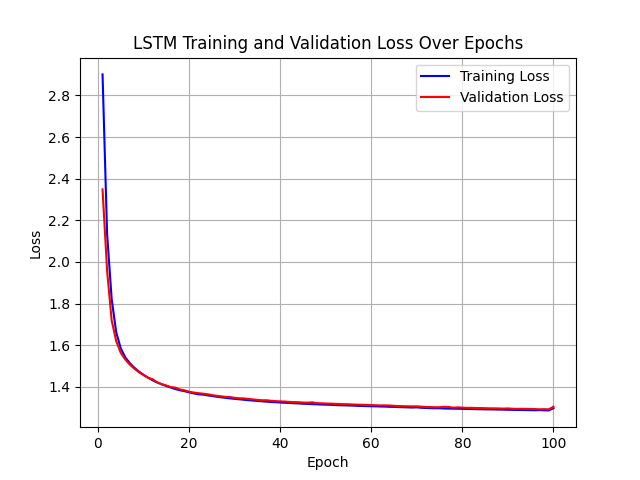}
        \caption{BPIC\_2019}
    \end{subfigure}
    \begin{subfigure}[b]{0.3\textwidth}
        \centering
        \includegraphics[width=\textwidth]{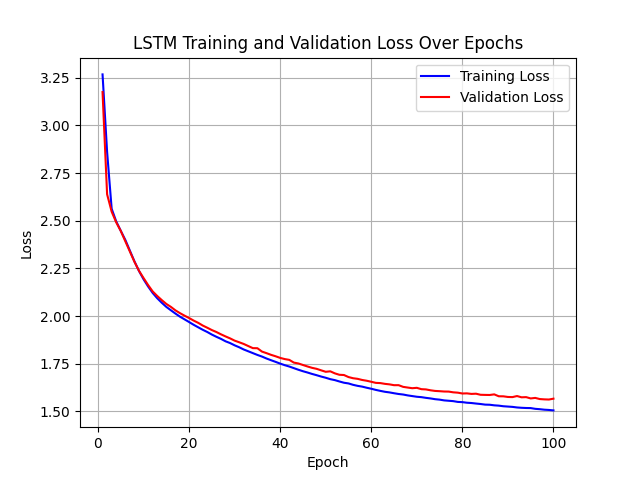}
        \caption{Sepsis\_Log}
    \end{subfigure}
    \begin{subfigure}[b]{0.3\textwidth}
        \centering
        \includegraphics[width=\textwidth]{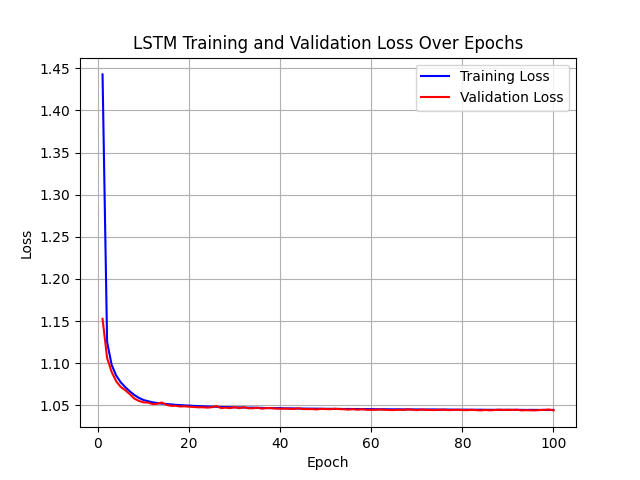}
        \caption{Traffic\_Fines}
    \end{subfigure}

    \caption{Training and validation loss using \textbf{multi-task LSTM} over epochs.}
    \label{fig:lstm_train_val_loss}
\end{figure}

In \Cref{fig:lstm_train_val_loss}, the LSTM model successfully converged on most datasets, indicating stable training performance. In addition to the LSTM architecture, a K-Nearest Neighbor (KNN) model was trained on the same data to support suffix replacement during evaluation. As previously discussed, the KNN model identifies the closest historical suffixes for a given prefix that are likely to result in a desired outcome. If the LSTM model predicts an undesired outcome, alternative suffixes retrieved by the KNN model can be applied. The suggested trace then provides insight into how far the alternative deviates from the ground truth suffix.

\begin{figure}[h!]
    \centering

    \begin{subfigure}[b]{0.32\textwidth}
        \centering
        \includegraphics[width=\textwidth]{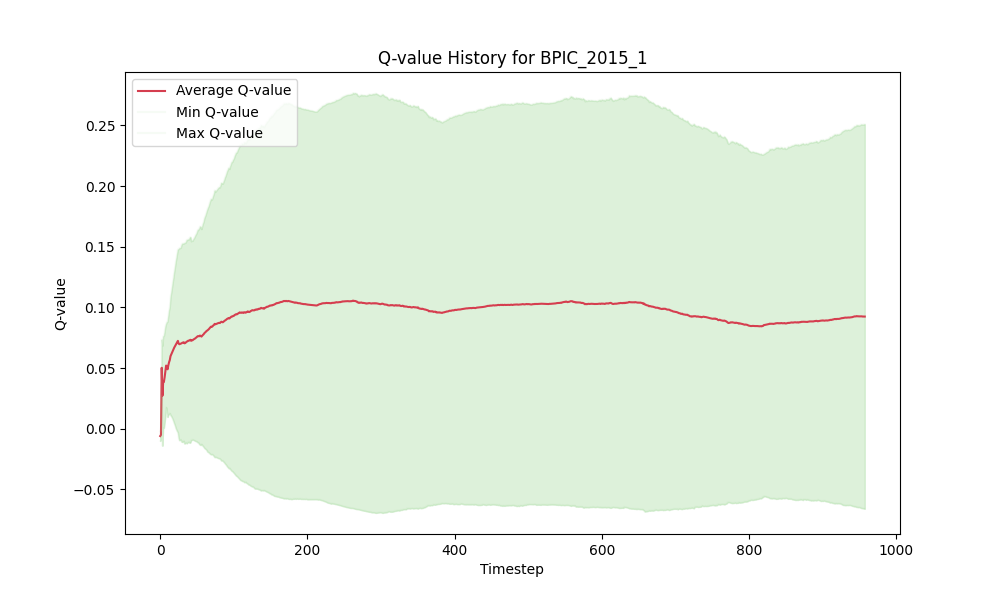}
        \caption{BPIC\_2015\_1}
    \end{subfigure}
    \begin{subfigure}[b]{0.32\textwidth}
        \centering
        \includegraphics[width=\textwidth]{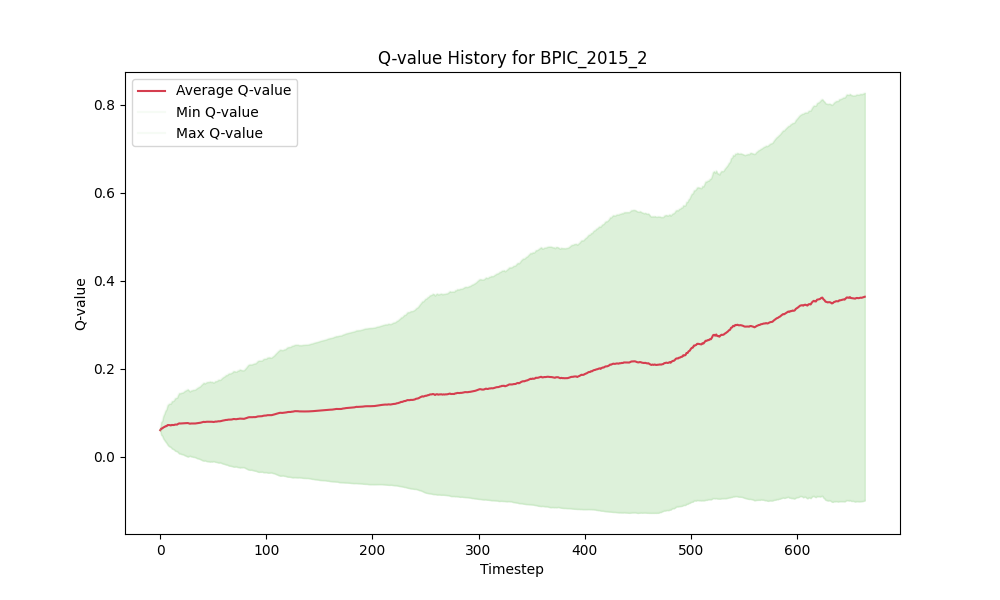}
        \caption{BPIC\_2015\_2}
    \end{subfigure}
    \begin{subfigure}[b]{0.32\textwidth}
        \centering
        \includegraphics[width=\textwidth]{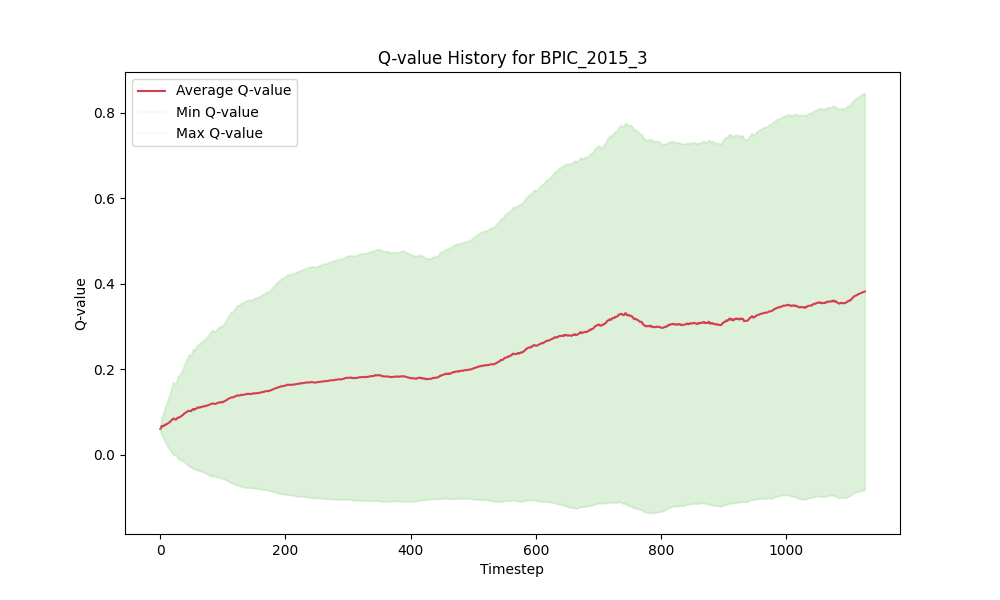}
        \caption{BPIC\_2015\_3}
    \end{subfigure}

    \vspace{0.3em}
    \begin{subfigure}[b]{0.32\textwidth}
        \centering
        \includegraphics[width=\textwidth]{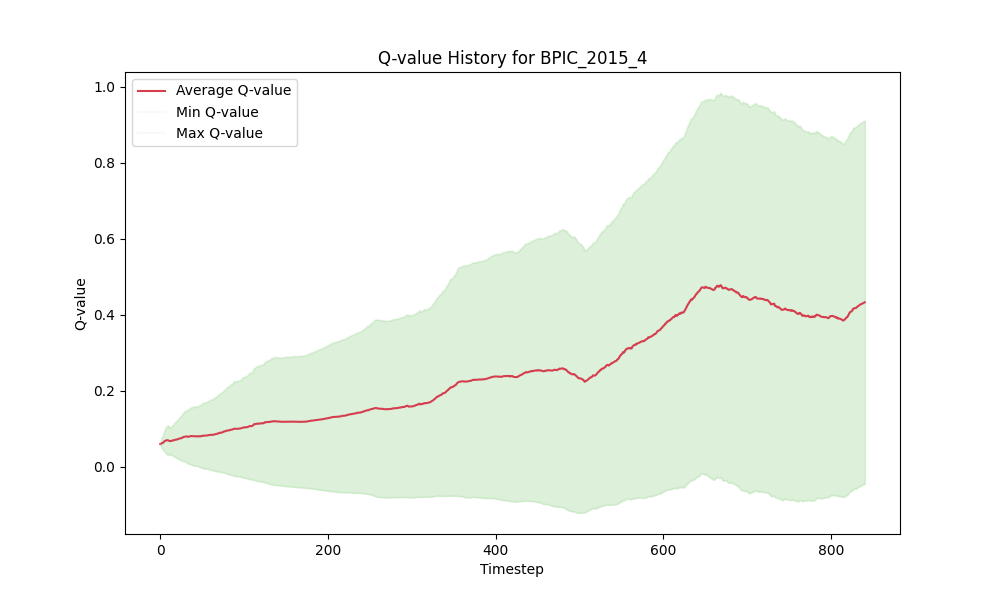}
        \caption{BPIC\_2015\_4}
    \end{subfigure}
    \begin{subfigure}[b]{0.32\textwidth}
        \centering
        \includegraphics[width=\textwidth]{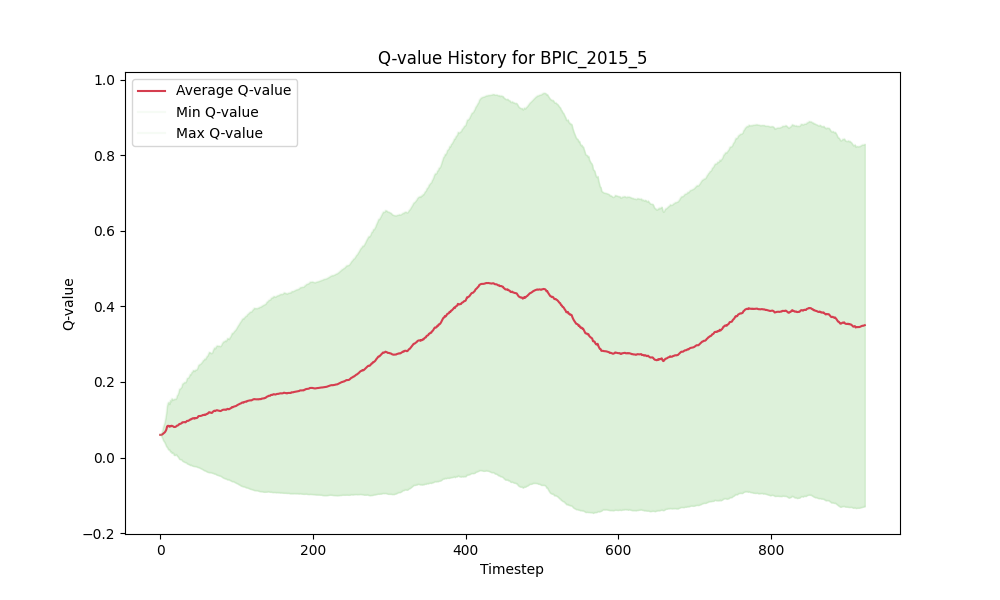}
        \caption{BPIC\_2015\_5}
    \end{subfigure}
    \begin{subfigure}[b]{0.32\textwidth}
        \centering
        \includegraphics[width=\textwidth]{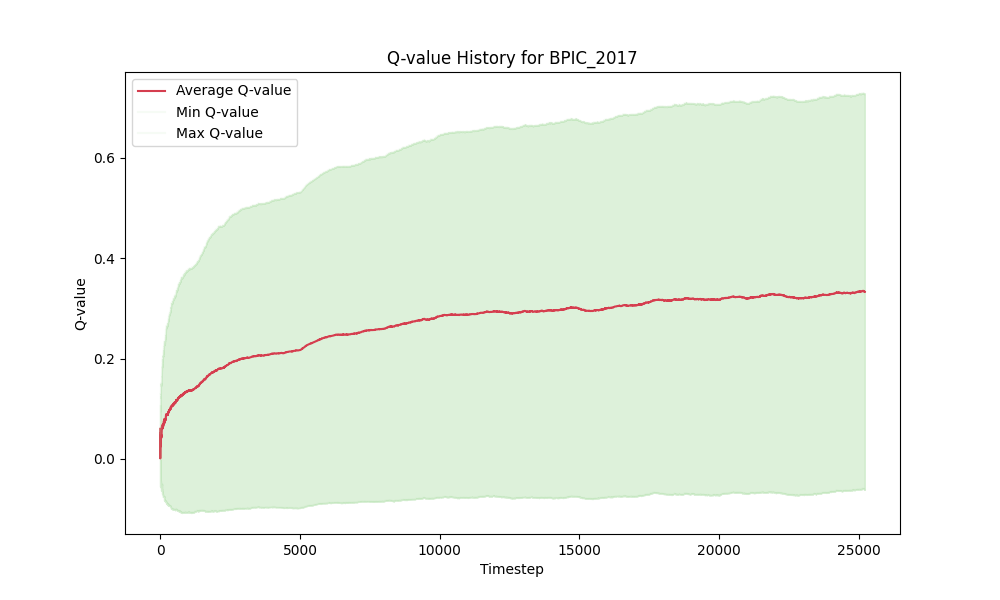}
        \caption{BPIC\_2017}
    \end{subfigure}

    \vspace{0.3em}
    \begin{subfigure}[b]{0.32\textwidth}
        \centering
        \includegraphics[width=\textwidth]{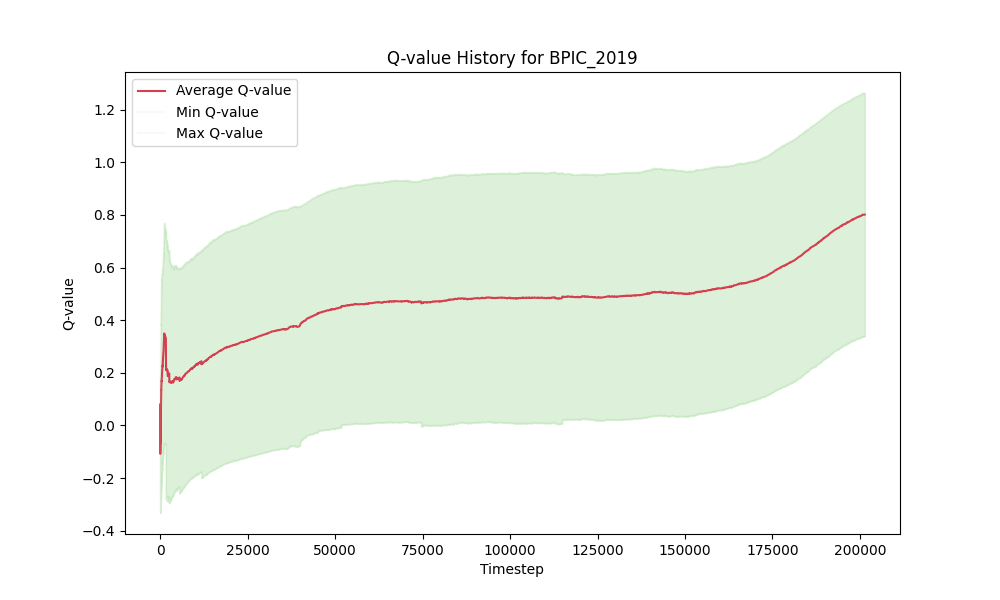}
        \caption{BPIC\_2019}
    \end{subfigure}
    \begin{subfigure}[b]{0.32\textwidth}
        \centering
        \includegraphics[width=\textwidth]{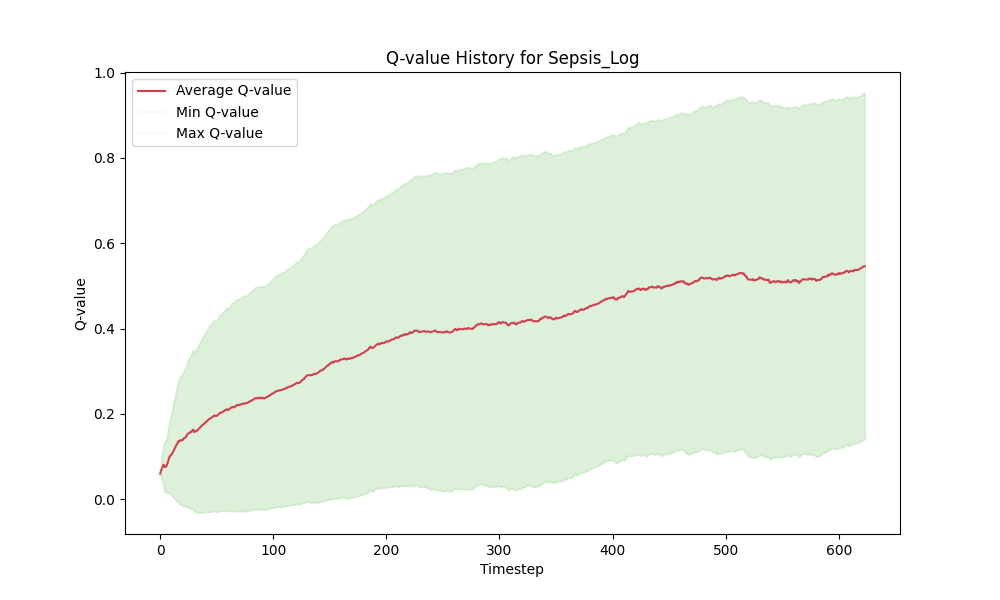}
        \caption{Sepsis\_Log}
    \end{subfigure}
    \begin{subfigure}[b]{0.32\textwidth}
        \centering
        \includegraphics[width=\textwidth]{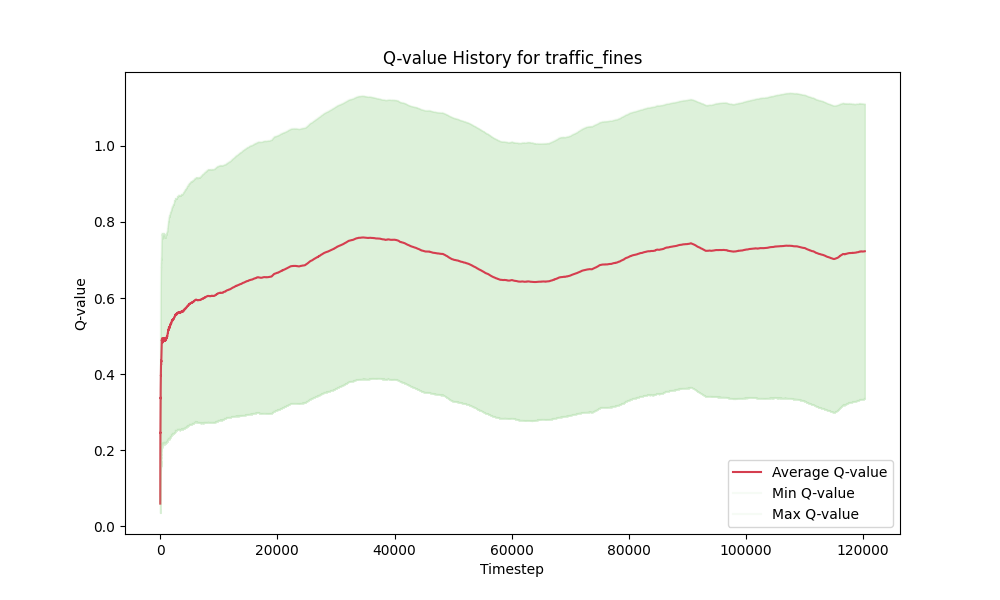}
        \caption{Traffic\_Fines}
    \end{subfigure}

    \caption{Q-value history across all datasets using \textbf{offline RL}.}
    \label{fig:_q_history}
\end{figure}

The Q-value progression for the offline RL method is presented in \Cref{fig:_q_history}. The results indicate that the model effectively learns from historical event logs, capturing meaningful patterns that support the optimal sequencing and prioritization of tasks, considering the process outcomes such as identifying patients likely to return to the emergency room within 28 days of discharge in the Sepsis dataset.

In the Sepsis case, the Markov Decision Process (MDP) was designed using a state-dependent reward function. If the patient returns to the emergency room (undesired outcome), each action is penalized proportionally to its position in the sequence. Later actions receive harsher penalties, highlighting potential misprioritization in earlier steps. If the outcome is favorable (i.e., the patient does not return), a fixed reward is assigned. This length-aware penalty structure enables the model to detect suboptimal decisions early in the trace and discourages inefficient activity ordering. The same reward shaping strategy was adapted for other datasets, based on their respective definitions of desired outcomes.

\Cref{fig:_q_history} demonstrates that the algorithm achieves stable convergence under the defined environment using fine-tuned hyperparameters. The availability of Q-values for each task within each state allows for the identification of optimal activity sequences within a trace. \Cref{tab:freq} presents a sample of these Q-values alongside the corresponding best actions in Sepsis case, illustrating how the learned policy prioritizes decisions based on the state-specific value estimations.

\begin{table}[ht]
\caption{Sample of a three-prefix trace (StateLength) and the best next activity in a Sepsis case}
\label{tab:freq}
\scriptsize
\resizebox{\textwidth}{!}{%
\begin{tabular}{lcccl}
\cline{1-4}
\textbf{State}                     & \textbf{StateLength} & \textbf{BestNextAction} & \textbf{Q} &  \\ \cline{1-4}
(ER Registration, ER Triage, ER Sepsis Triage) & 3                    & CRP             & 1.089389	   &  \\
(ER Registration, ER Triage, IV Liquid)	      & 3                    & ER Sepsis Triage	             & 0.117600	   &  \\
(ER Registration, CRP, Leucocytes)      & 3                  & LacticAcid              & 0.276705	    \\ 
\cline{1-4}

\end{tabular}
}
\end{table}

Given the limited size of available datasets, data augmentation is critical for supporting model convergence. To address this, we employed an augmentation strategy that increased the number of episodes per dataset and introduced 10\% random noise to improve variability.

\begin{figure}[h!]
    \centering
   
    \begin{subfigure}[b]{0.32\textwidth}
        \centering
        \includegraphics[width=\textwidth]{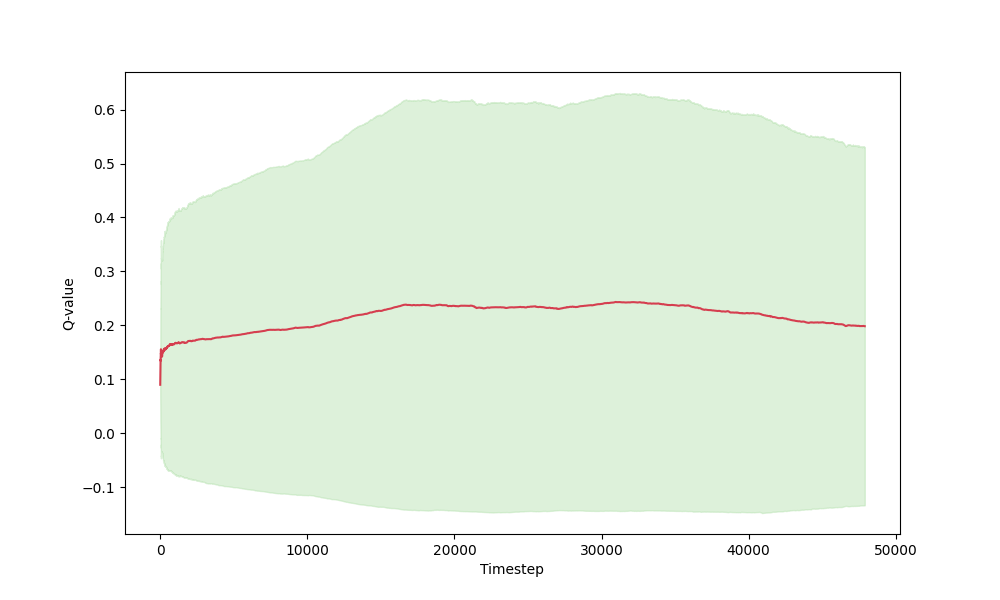}
        \caption{BPIC\_2015\_1}
    \end{subfigure}
    \begin{subfigure}[b]{0.32\textwidth}
        \centering
        \includegraphics[width=\textwidth]{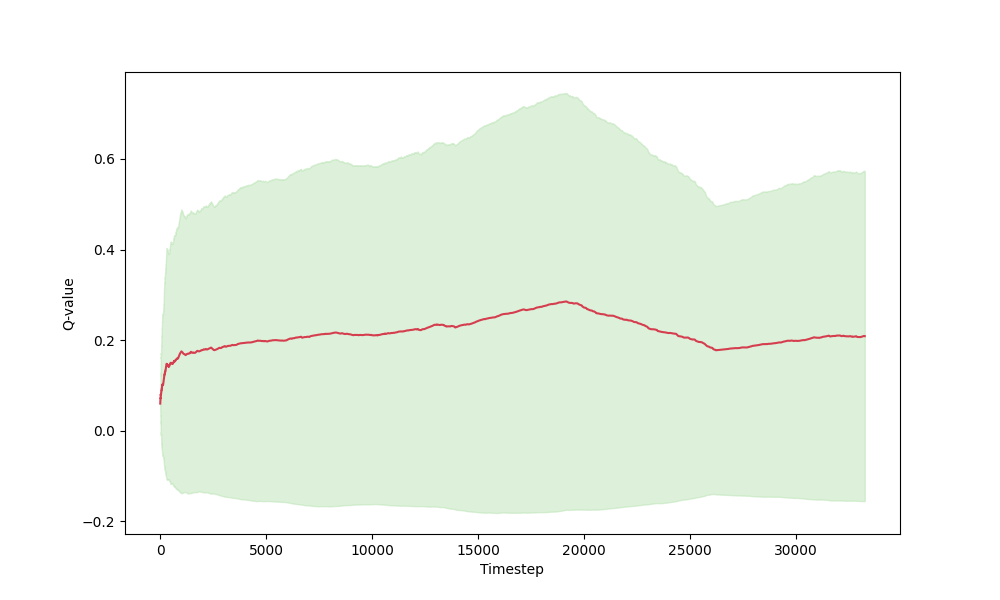}
        \caption{BPIC\_2015\_2}
    \end{subfigure}
    \begin{subfigure}[b]{0.32\textwidth}
        \centering
        \includegraphics[width=\textwidth]{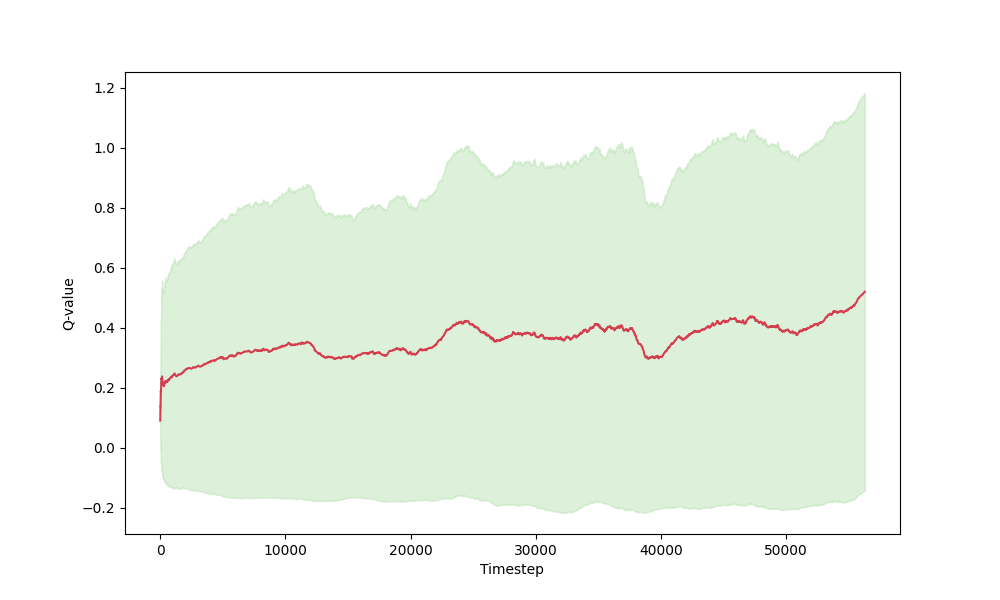}
        \caption{BPIC\_2015\_3}
    \end{subfigure}

    \vspace{0.1em}
    \begin{subfigure}[b]{0.32\textwidth}
        \centering
        \includegraphics[width=\textwidth]{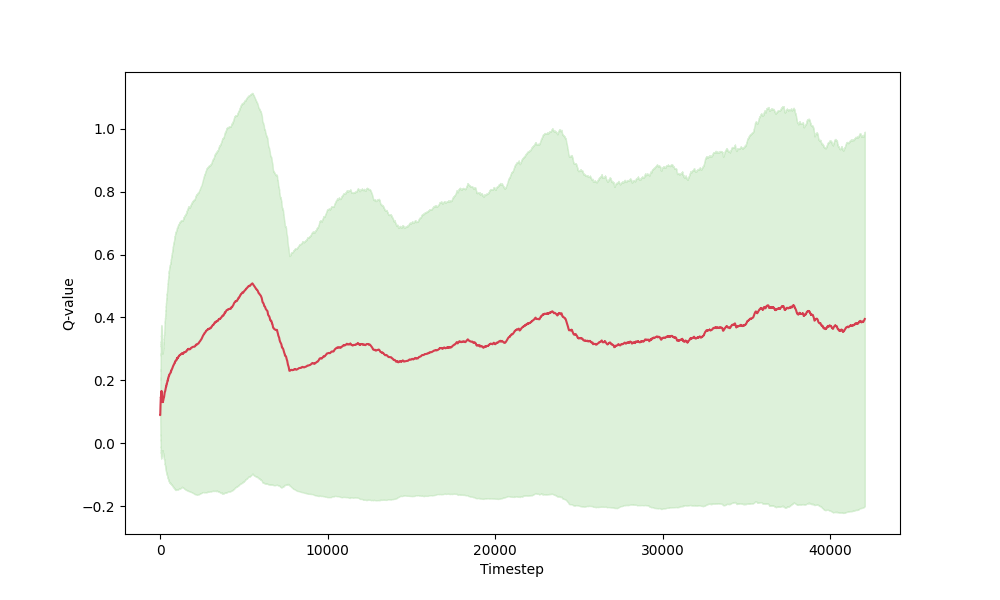}
        \caption{BPIC\_2015\_4}
    \end{subfigure}
    \begin{subfigure}[b]{0.32\textwidth}
        \centering
        \includegraphics[width=\textwidth]{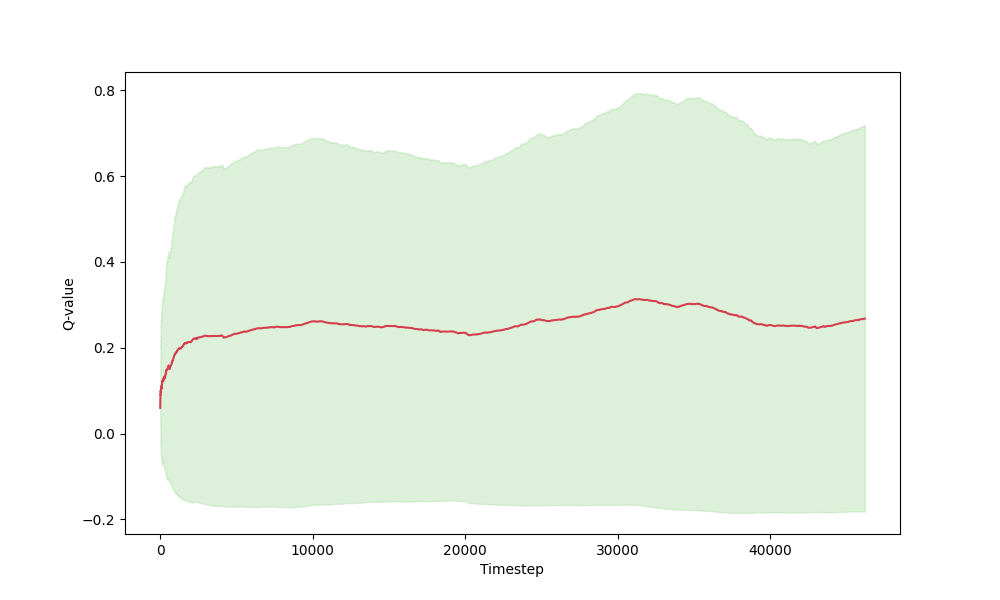}
        \caption{BPIC\_2015\_5}
    \end{subfigure}
    \begin{subfigure}[b]{0.32\textwidth}
        \centering
        \includegraphics[width=\textwidth]{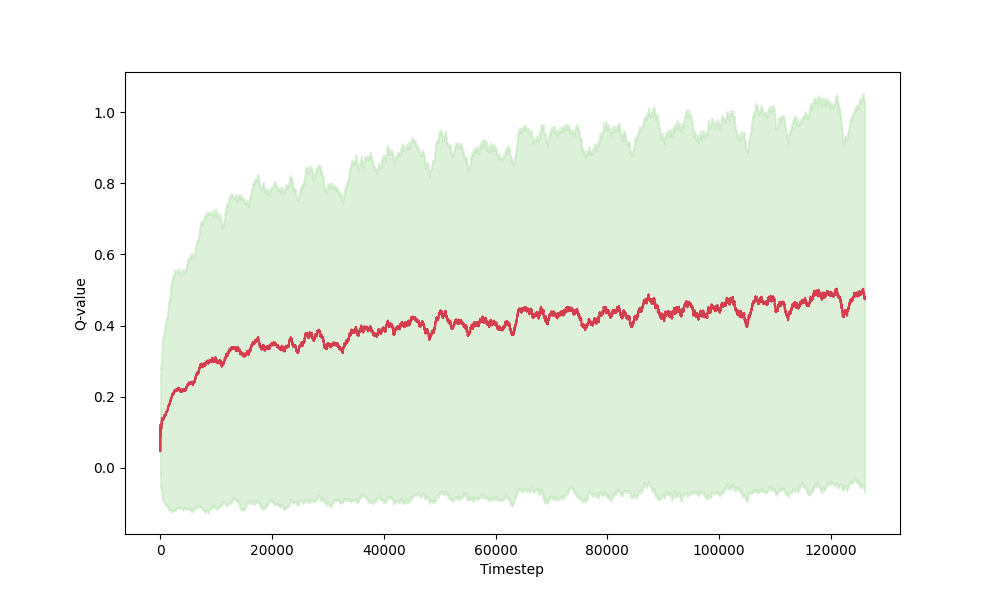}
        \caption{BPIC\_2017}
    \end{subfigure}

    \vspace{0.1em}
    \begin{subfigure}[b]{0.32\textwidth}
        \centering
        \includegraphics[width=\textwidth]{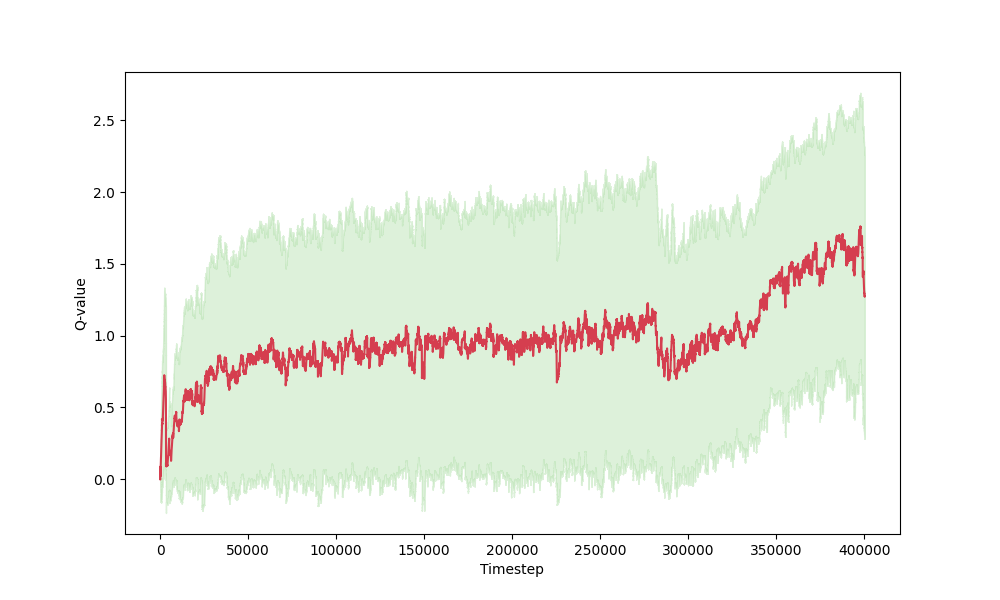}
        \caption{BPIC\_2019}
    \end{subfigure}
    \begin{subfigure}[b]{0.32\textwidth}
        \centering
        \includegraphics[width=\textwidth]{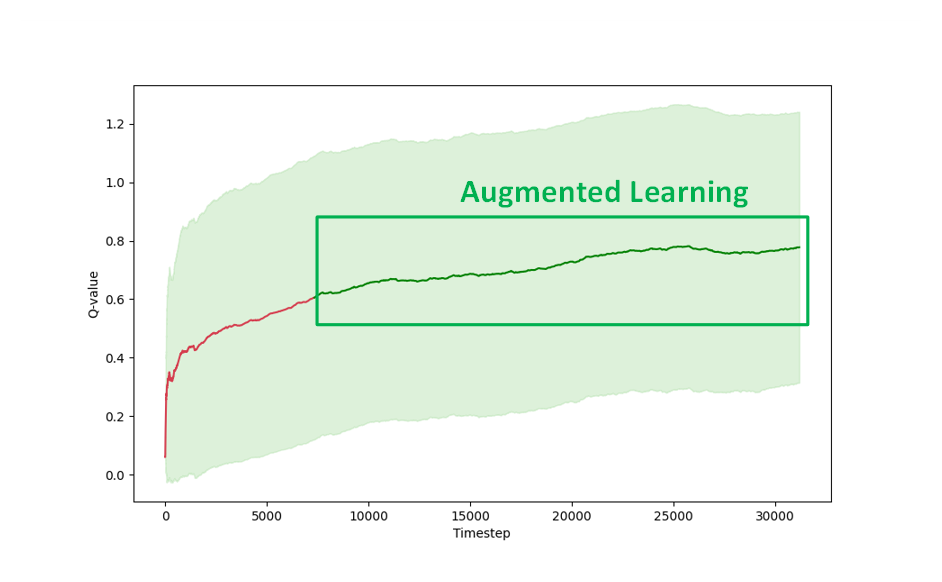}
        \caption{Sepsis\_Log}
        \label{fig:forlaps_sepsis}
    \end{subfigure}
    \begin{subfigure}[b]{0.32\textwidth}
        \centering
        \includegraphics[width=\textwidth]{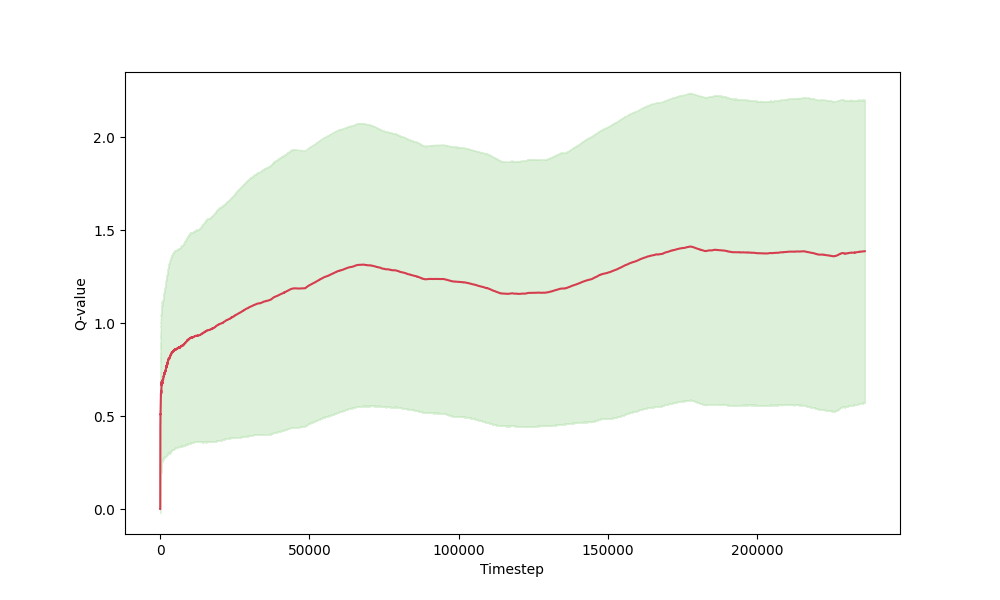}
        \caption{Traffic\_Fines}
    \end{subfigure}

    \caption{Q-value history across all datasets using \textbf{FORLAPS}.}
    \label{fig:forlaps_q_history}
\end{figure}

Fine-tuning the pre-trained offline RL model, as depicted in \Cref{fig:forlaps_q_history}, initially resulted in a temporary decline in the training Q-value across most cases. As an example, in \Cref{fig:forlaps_sepsis}, the green line represents the learning steps for the augmented dataset, while the red line corresponds to the original dataset. Over time, the model achieved convergence, culminating in a substantial improvement in the Q-mean.

As a result, Offline RL is characterized by fast initial learning progress, but it tends to plateau early due to limited exposure to diverse trajectories. In contrast, FORLAPS demonstrates sustained improvement, enabled by augmentation-based fine-tuning. This suggests that a hybrid strategy, starting with offline training and continuing with data augmentation, can enhance policy quality without requiring real-time environment interaction. In effect, augmented datasets simulate the benefits of continued exploration, allowing the agent to discover improved behavior over time. Had training continued longer, we expect the FORLAPS model to continue improving, whereas purely offline agents would likely stagnate.

Offline Q-learning with fine-tuning demonstrated significantly superior performance compared to counterparts. Notably, it outperformed the Offline RL agent by 28 percentage points on the Sepsis dataset and 33 percentage points on the BPIC\_2017 dataset. On the Traffic Fines dataset, Q-learning with fine-tuning achieved improvements in the agent’s final performance of up to 19 percentage points. These results highlight the method’s effectiveness in optimizing decision-making policies, which can be attributed to its ability to achieve a higher learning rate across traces, enabling more efficient identification of optimal activity sequences.

The performance improvement achieved by FORLAPS varies across datasets, largely due to differences in trace sparsity and state-space coverage. For instance, in BPIC\_2015s, the combination of extremely high activity diversity (over 300 tasks) and a relatively low number of unique states (10K) results in a sparse and underexplored state-transition space. Many theoretically valid sequences of tasks are either rarely observed or entirely absent in the log. As a result, the RL agent is frequently exposed to states where numerous actions are technically possible, but the data only shows a limited subset. This gap between possible and observed transitions leads to high uncertainty in Q-value estimation, hindering the model’s ability to generalize and resulting in unstable learning dynamics, as evidenced by persistent fluctuations in Q-values during training (\Cref{fig:forlaps_q_history}).

In contrast, BPIC\_2017 and BPIC\_2019 present more favorable environments for offline RL. BPIC\_2017 contains 26 unique activities but covers over 66K unique states, while BPIC\_2019 includes 41 activities and nearly 48K unique states. Despite their complexity and the potential for high variability, these datasets provide richer state-transition coverage, allowing the RL agent to observe each action in a variety of contexts. This density enables more stable reward estimation and policy convergence. Although training curves for these datasets may exhibit some noise—partly due to their size and the variability introduced by augmentation—the models ultimately achieve better convergence and improved performance.

In essence, FORLAPS performs best in settings where the state space is densely covered relative to the action space, as in BPIC\_2017 and BPIC\_2019. These logs offer a robust foundation of observed transitions, supporting reliable learning of optimal policies. Conversely, datasets like BPIC\_2015s, with a disproportionately large action space and limited scenario realization, pose significant challenges for Q-learning and require further exploration or domain-specific adaptations to improve performance.

\subsection{Validation}

In this experiment, we conducted a comparative evaluation against an LSTM-based and Offline RL baseline, focusing on two key aspects: state-dependent reward shaping and process data augmentation. The first objective was to assess how a reinforcement learning (RL) agent—trained with a reward function sensitive to the position and impact of each action—performs in comparison to LSTM when predicting sequences leading to continuous outcomes. The second objective involved evaluating the effectiveness of Fine-Tuned Offline Reinforcement Learning Augmented Process Sequence Optimization (FORLAPS) against standard offline RL, to demonstrate how augmenting the dataset can enhance the agent's experience and improve generalization, especially in cases with limited trace availability.
One of the critical challenges in evaluation arises from the absence of real-life event logs that encompass all possible scenarios, making it difficult to comprehensively assess model performance. To address this, we designed an experiment where the dataset was split into training and testing sets (80-20\%), and different methods were compared based on the similarity between the recommended path and the actual path for desired and undesired outcomes. The evaluation was performed using the average Damerau-Levenshtein distance, a metric that quantifies the minimum number of operations (insertions, deletions, or substitutions) required to transform one process trace into another. A lower Damerau-Levenshtein distance indicates a higher similarity between the traces.

\begin{figure}[htbp]
    \centering
    \begin{subfigure}[b]{0.48\textwidth}
        \centering
        \includegraphics[width=\textwidth]{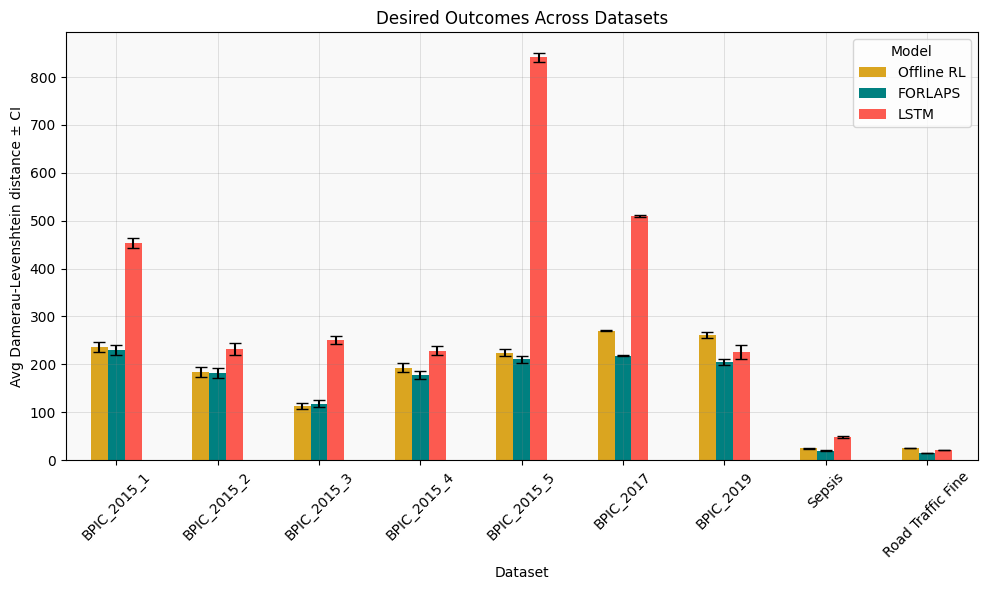}
        \caption{Desired outcome}
    \end{subfigure}
    \hspace{0.02\textwidth}
    \begin{subfigure}[b]{0.48\textwidth}
        \centering
        \includegraphics[width=\textwidth]{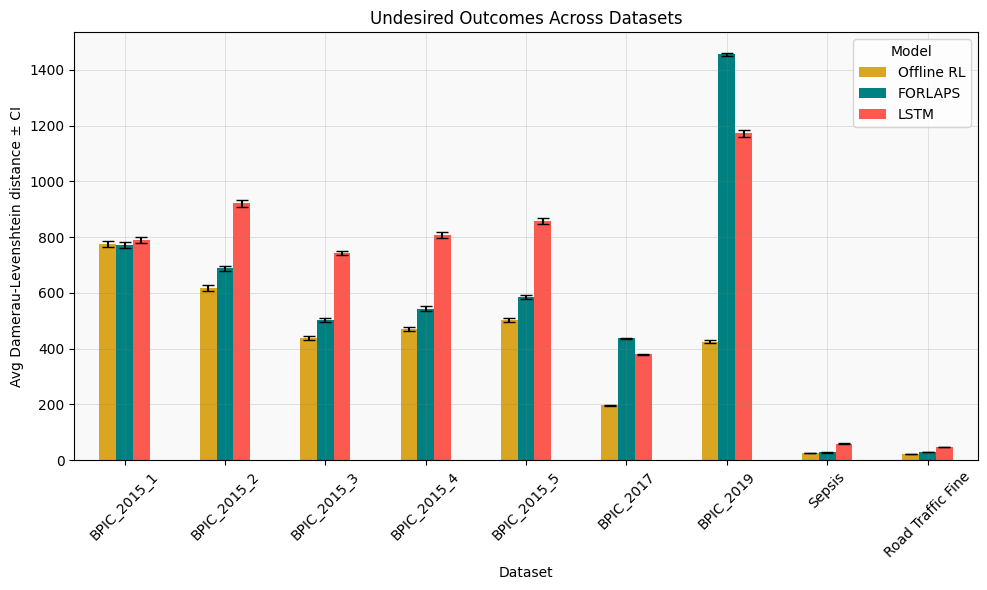}
        \caption{Undesired outcome}
    \end{subfigure}
    \caption{Mean Damerau--Levenshtein distance with 95\% confidence intervals for recommended paths corresponding to desired and undesired outcomes across test set traces.}
    \label{distance_eval}
\end{figure}

In \Cref{distance_eval}, we compare the suggested routes with the ground truth for both positive and negative results, reporting the average distances and the related 95\% confidence intervals. Greater accuracy in reproducing ideal process traces, even for cases not in the training set, is shown by closer alignment with the desired route. A greater departure from traces that result in unfavourable outcomes, on the other hand, indicates the model's capacity to recognise and steer clear of less-than-ideal routes, which is an essential component of assessing the overall efficacy of the suggested techniques. Dissimilarity by itself, however, does not prove that the suggested course is the most effective one. Rather, it shows if the procedure prevents the same incorrect patterns from happening again.

We observed that augmentations affect the learning performance of different datasets in varying ways. For BPIC\_2015, despite the poor performance in both offline reinforcement learning and FORLAPS, the latter outperformed LSTM by a significant margin in terms of closeness to the desired outcome. This can be attributed to the adaptive capabilities of the FORLAPS approach over the LSTM and KNN models, especially for highly volatile cases. Similarly, in BPIC\_2017, RL approaches were able to suggest alternatives closer to the desired outcome, while LSTM took a more conservative approach in detecting and predicting negative outcomes, recommending further traces. In contrast, BPIC\_2019 showed that LSTM slightly outperformed offline RL, though FORLAPS achieved higher average performance, surpassing offline RL and LSTM by 9\%. In the Sepsis and Traffic Fines datasets, all methods exhibited similar behavior, with FORLAPS leading, followed by offline RL and LSTM.
Overall, FORLAPS demonstrated that, across most datasets, it could provide the next best activities within a trace with greater efficiency and stability. In data sets such as BPIC\_2017 and BPIC\_2019, considered the most appropriate for assessing and evaluating the method, the results revealed that the model possesses superior exploratory capabilities and can adapt to data sets where certain patterns were previously unobserved, as it explored various possible traces and learned accordingly.

\begin{figure}[htbp]
    
    \centering
    \begin{subfigure}[b]{0.3\textwidth}
        \centering
        \includegraphics[width=\textwidth]{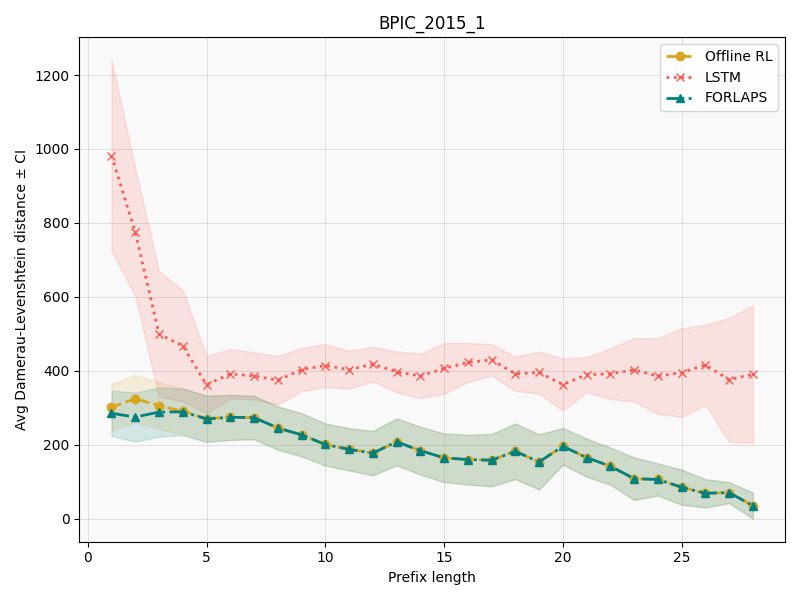}
        \caption{BPIC\_2015\_1}
    \end{subfigure}
    \hfill
    \begin{subfigure}[b]{0.3\textwidth}
        \centering
        \includegraphics[width=\textwidth]{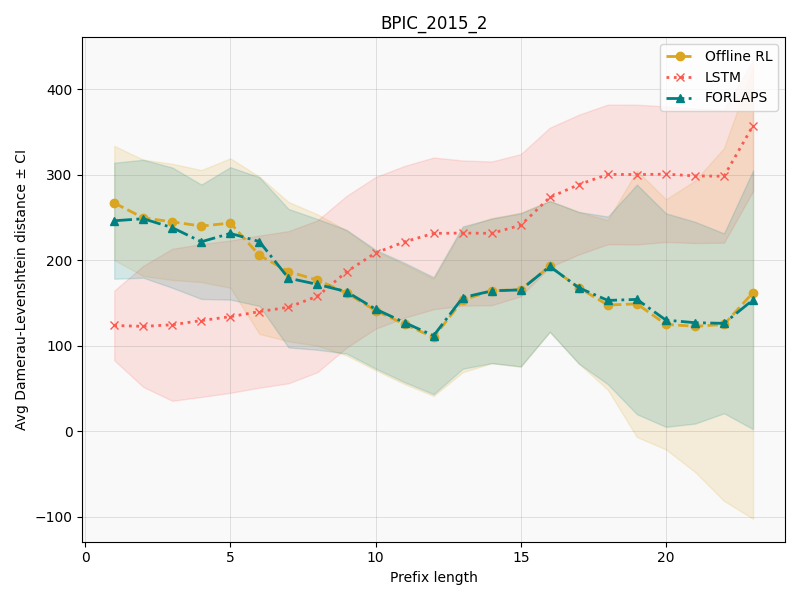}
        \caption{BPIC\_2015\_2}
    \end{subfigure}
    \hfill
    \begin{subfigure}[b]{0.3\textwidth}
        \centering
        \includegraphics[width=\textwidth]{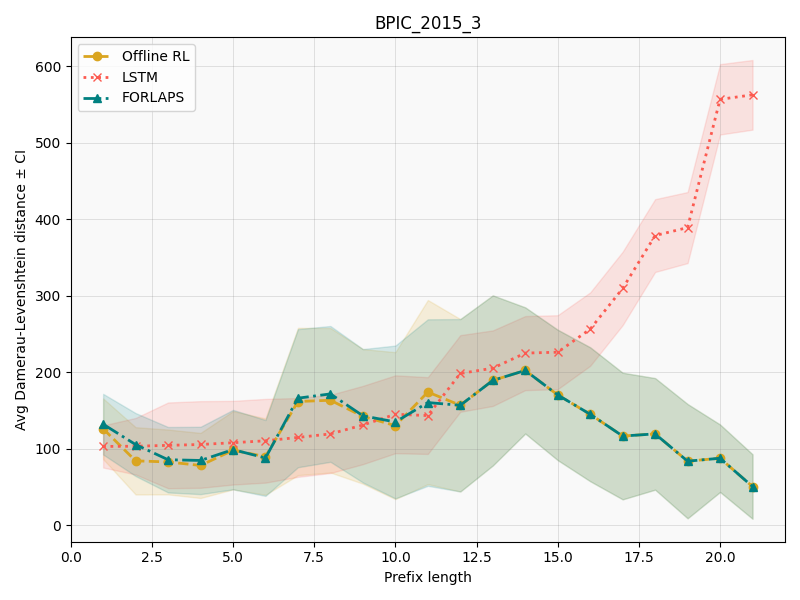}
        \caption{BPIC\_2015\_3}
    \end{subfigure}

    \vspace{0.5em}

    \begin{subfigure}[b]{0.3\textwidth}
        \centering
        \includegraphics[width=\textwidth]{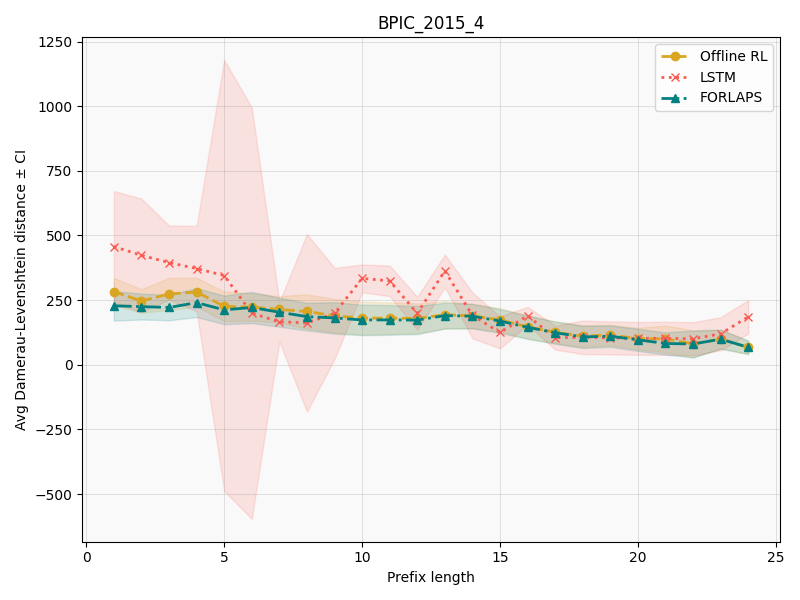}
        \caption{BPIC\_2015\_4}
    \end{subfigure}
    \hfill
    \begin{subfigure}[b]{0.3\textwidth}
        \centering
        \includegraphics[width=\textwidth]{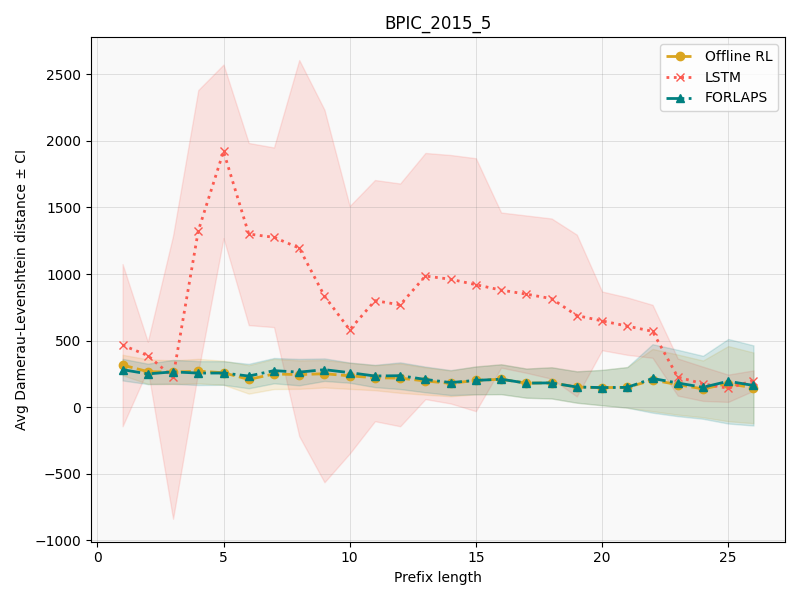}
        \caption{BPIC\_2015\_5}
    \end{subfigure}
    \hfill
    \begin{subfigure}[b]{0.3\textwidth}
        \centering
        \includegraphics[width=\textwidth]{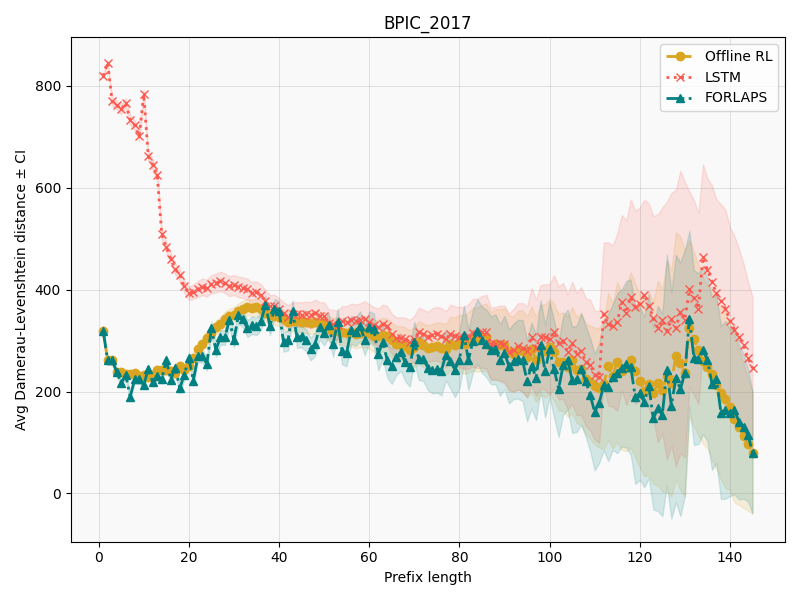}
        \caption{BPIC\_2017}
    \end{subfigure}

    \vspace{0.5em}

    \begin{subfigure}[b]{0.3\textwidth}
        \centering
        \includegraphics[width=\textwidth]{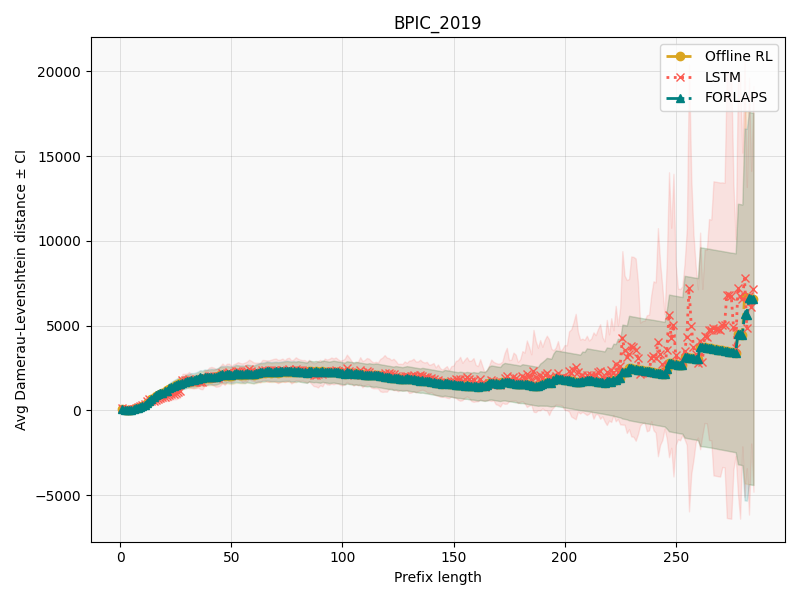}
        \caption{BPIC\_2019}
    \end{subfigure}
    \hfill
    \begin{subfigure}[b]{0.3\textwidth}
        \centering
        \includegraphics[width=\textwidth]{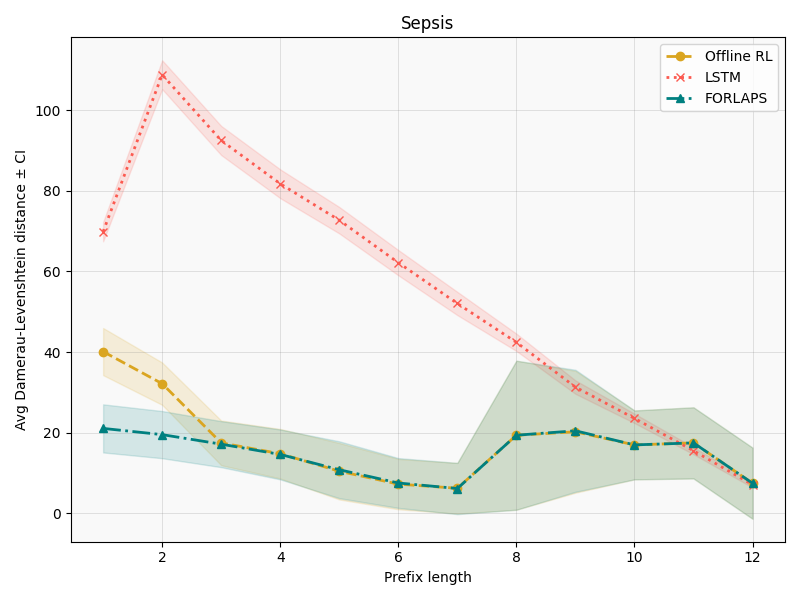}
        \caption{Sepsis\_1}
    \end{subfigure}
    \hfill
    \begin{subfigure}[b]{0.3\textwidth}
        \centering
        \includegraphics[width=\textwidth]{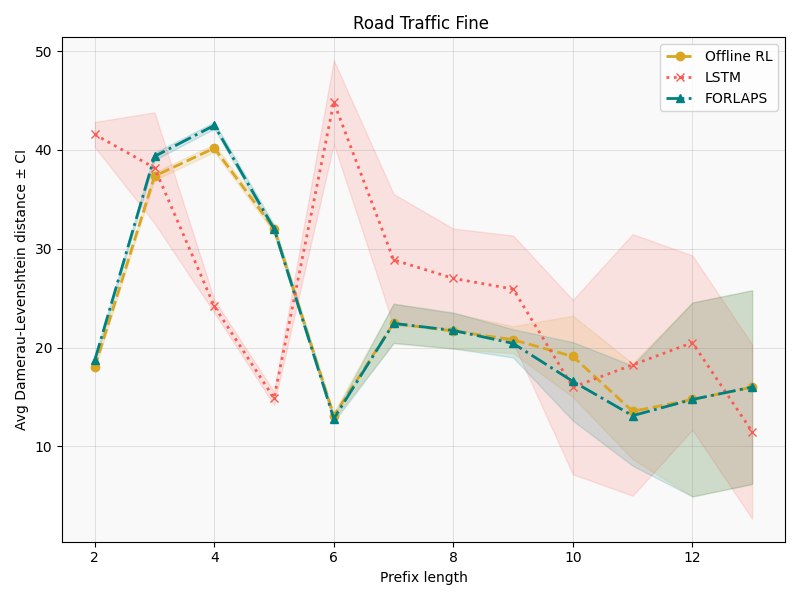}
        \caption{Road\_Traffic\_fine}
    \end{subfigure}

    \caption{Mean Damerau--Levenshtein distance with 95\% confidence intervals for desired outcomes across test set traces at varying prefix lengths.}
    \label{fig:distance_prefix_all}
\end{figure}

One of the key challenges in these works is the development of generative models based on the environment and outcomes or probabilistic models. These models act as simulated environments to assess the effectiveness of various approaches, a crucial factor given the high cost of data collection in real-world scenarios. The more closely the model can replicate real-world behavior, the more effectively the agent can learn and adapt. A second challenge, as highlighted by our experiments on certain cases such as BPIC\_2015s, is that data augmentation techniques may not provide the expected benefits. This could be due to their inappropriateness for cases with a large number of distinct activities. Therefore, in addition to the significance of the number of activities and prefixes in process traces, most process data suffers from data imbalance, with rare cases of violations in some instances.

We conducted a complementary analysis in which we assessed model performance over different prefix lengths. In \Cref{fig:distance_prefix_all}, the Damerau–Levenshtein distance was presented as a line chart for each prefix size, with shaded areas signifying confidence intervals.  The findings demonstrate that RL-based methods consistently performed better than the LSTM baseline, especially for longer prefixes. The penalisation process of RL's reward function, which imposes progressively harsher penalties on longer, inefficient traces, is responsible for this improvement.  Furthermore, FORLAPS's augmentation component increased stability for all prefix lengths, which improved the model's capacity to generalise and sustain performance even when dealing with long traces.

\section{Conclusions, limitations and future works}
Our novel AI-driven approach tackles the challenge of activity sequence optimization in business process management by introducing a data-centric framework that integrates process-aware data augmentation with reinforcement learning. FORLAPS, a five-step framework, enhances the stability and effectiveness of conventional offline RL models by leveraging event log augmentation and fine-tuning strategies, ultimately improving policy learning and decision-making in complex, real-life business scenarios.

The proposed FORLAPS framework demonstrates strong applicability in industrial settings with highly parallel process structures, where the ability to distinguish the impact of activity ordering is essential. Applied to a real-world event log, FORLAPS achieved notable improvements in process performance, including a 31\% reduction in resource time and a 23\% decrease in total process duration. FORLAPS continuously outperformed conventional models, such as LSTM and permutation feature importance (PFI) baselines, in terms of the specified key performance indicators (KPIs). The intrinsic drawbacks of offline reinforcement learning, including learning instability and restricted generalisation, are successfully addressed by FORLAPS, a hybrid technique that combines offline pre-training with online fine-tuning. Notably, offline Q-learning with fine-tuning demonstrated its resilience and adaptability in a variety of process environments by achieving performance improvements of up to 19 percentage points on the Traffic Fines dataset and 28 and 33 percentage points on the Sepsis and BPIC\_2017 datasets, respectively. Furthermore, our analysis revealed that a key determinant of model performance lies in the degree of uncertainty and sparsity within the process flow. This can be effectively assessed by examining the relationship between the number of distinct activities and the coverage of unique process states. These structural indicators are essential for evaluating the stability, complexity, and suitability of event logs for reinforcement learning-based prescriptive monitoring, and offer practical guidance for dataset selection and preparation in real-world applications.

To evaluate the agent’s ability to generalize and make effective prescriptive decisions, we implemented a complementary method based on Damerau–Levenshtein distance. The results confirmed that FORLAPS is capable of replicating desirable process behaviors while effectively avoiding negative execution patterns. This was validated across nine public datasets through two experiments focused on trace-level and prefix-level evaluations, emphasizing the value of state-dependent reward shaping in guiding the model to consider both the current context and downstream consequences. Overall, our findings highlight the effectiveness and robustness of FORLAPS in optimizing decision-making across diverse and complex process environments, offering a scalable and generalizable solution for prescriptive process monitoring.

In our experiments, to showcase the generalizability of the proposed method across various industries, including healthcare, business, manufacturing, and government, we incorporated diverse reward structures based on process outcomes from a case study and nine additional datasets. These variations significantly influenced model convergence and robustness.

While the results are promising, reinforcement learning (RL) approaches still require careful customization of the Markov Decision Process (MDP) to ensure effective policy learning, as well as the definition of business rules for process data augmentation. To ensure convergence and effectiveness in prescriptive process monitoring, it is essential that the augmented data maintains both conformity and integrity. They ensure that the augmented data adheres to established business rules and process constraints, preventing the introduction of unrealistic scenarios that could mislead the learning process and guarantee that the augmented data accurately reflects the underlying process semantics, preserving the relationships and dependencies inherent in the original data. Without these safeguards, there is a risk of introducing biases or inconsistencies that could degrade the performance and reliability of the RL model. Moreover, there remains a substantial opportunity to develop more robust and autonomous RL frameworks that can identify optimal paths across various domains with minimal manual intervention. However, excessive reliance on predefined business rules or the potential biases introduced by augmented data could limit the scalability and adaptability of such frameworks. 

Future work could explore the integration of automated reward shaping and advanced augmentation techniques to improve the model's convergence stability and adaptability, ensuring more consistent and reliable performance across diverse scenarios. Incorporating these advancements could enhance the reliability, adaptability, and scalability of RL-based prescriptive process monitoring systems, reducing the need for manual intervention and making them better suited for a wide range of real-world applications.

\section{ACKNOWLEDGMENT} We acknowledge the support of the Natural Sciences and Engineering Research Council of Canada (NSERC), [funding reference number ALLRP 561264 - 21]. 
Cette recherche a été financée par le Conseil de recherches en sciences naturelles et en génie du Canada (CRSNG), [numéro de référence ALLRP 561264 - 21]

\bibliographystyle{cas-model2-names}
\bibliography{main}

\end{document}